\documentclass{article}

\usepackage{iclr2027_conference,times}

\usepackage{amsmath,amsfonts,bm}

\def\eqref#1{equation~\ref{#1}}

\def\1{\bm{1}}

\DeclareMathAlphabet{\mathsfit}{\encodingdefault}{\sfdefault}{m}{sl}
\SetMathAlphabet{\mathsfit}{bold}{\encodingdefault}{\sfdefault}{bx}{n}

\usepackage{hyperref}
\usepackage{url}
\usepackage{booktabs,graphicx}
\usepackage{adjustbox}
\usepackage{multirow}

\title{\textsc{Unitas}: A 3D-Native World Action Model for Embodied Manipulation}

\author{Ruixiang Wang$^{1,2}$\textsuperscript{*} \quad Yongyi Su$^{3}$\textsuperscript{*\textdagger} \quad Wenlve Zhou$^{3}$ \quad Bo Yue$^{1,2}$ \quad Hengyan Liu$^{1,2}$ \quad Dekun Lu$^{2}$ \\
\bf Yuxin Tian$^{1,2}$ \quad Yihan Fang$^{1,2}$ \quad Zerui Wu$^{2}$ \quad Xing Hu$^{2}$ \quad Jietao Chen$^{2}$ \quad Yong Guo$^{2}$ \\
\bf Ziyan He$^{2}$ \quad Junbin Yuan$^{4}$ \quad Guiliang Liu$^{1}$ \quad Xiaofen Xing$^{5}$ \quad Kui Jia$^{1,2}$\\
\\
$^1$ The Chinese University of Hong Kong, Shenzhen \quad
$^2$ DexForce Technology Co., Ltd.\\
$^3$ Foshan University\quad
$^4$ Sun Yat-sen University \quad
$^5$ South China University of Technology\\
\\
\textsuperscript{*} Equal contribution. \quad
\textsuperscript{\textdagger} Correspondence to \textless\href{mailto:suyongyi@fosu.edu.cn}{suyongyi@fosu.edu.cn}\textgreater.
}

\iclrfinalcopy

\newcommand{\method}{\textsc{Unitas}}

\begin{document}

\maketitle


\begin{abstract}
World action models (WAMs) aim to answer a coupled physical question: given a task instruction, what motion should the robot execute, and how will that motion change the surrounding world? Most existing WAMs build on pretrained video generators and represent world evolution through images or visual latents. Robotic interaction, however, takes place in metric three-dimensional space, while images are view-dependent projections whose pixel distances do not directly encode physical distances.
We introduce \method{}, to our knowledge the first 3D-native world action model that unifies observations, actions, and scene dynamics in a shared metric 3D frame within each interaction, using a common representation across robot embodiments and human hands. Action flow represents human hands and robot grippers as 3D point trajectories, while scene flow describes scene-point displacements conditioned on these trajectories. World-aligned 3D positional embeddings ground visual tokens with or without depth input, and a physical-time trajectory tokenizer encodes each point trajectory as one token anchored at its current 3D position. This interface supports both direct action execution and
action-conditioned scene prediction.
With 1.7B parameters, \method{} achieves the best action-conditioned
scene prediction on RoboTwin among the compared methods,
with up to 49\% lower displacement errors than PointWorld,
and state-of-the-art manipulation success, including 99.8\%
on LIBERO and an average of 85\%
across real-world tasks. The code is available at \url{https://github.com/DexForce/UNITAS}.

\end{abstract}

\vspace{-0.2cm}
\section{Introduction}
\vspace{-0.1cm}

\label{sec:introduction}

World Action Models (WAMs) aim to answer a coupled physical question:
\emph{given a task instruction, what motion should the robot execute,
and how will that motion change the surrounding world?}
Most existing WAMs couple action learning with prediction of future
images or visual latents, increasingly building on pretrained video models
\citep{yang2023unisim,li2025uva,zhu2025uwm,cen2025worldvla,
bi2025motus,ye2026dreamzero,yuan2026fastwam,zhou2026tau0wm}.
Robotic interaction, however, takes place in metric three-dimensional
space, and an image is only a view-dependent projection of that space.

The same robot motion, object displacement, or contact event can produce
different pixel changes across views, and equal pixel displacements
need not correspond to equal physical distances
\citep{ze2024threediffusion,qu2025spatialvla}.
Differences in camera intrinsics and camera-to-robot poses further
make this mapping vary across embodiments.
Evaluations of video-based WAMs report sensitivity to viewpoint and
robot-state changes \citep{zhang2026wamgeneralization}, and a
photorealistic prediction may still misrepresent object displacement,
deform rigid geometry, or allow static regions to drift
\citep{huang2026pointworld}.
Such visual representations leave metric motion and contact relationships
implicit for spatial reasoning, prediction, and control.

Recent methods incorporate depth, optical flow, scene flow, or
reconstructed 4D geometry into prediction
\citep{zhen2025tesseract,guo2026xwam,zhao2026rynnworld,wang2026lamp}
or ground action generation in explicit 3D structure
\citep{shridhar2023peract,ze2024threediffusion,zhen2024threedvla,
qu2025spatialvla}.
Yet geometry often serves as an input cue or auxiliary prediction target,
rather than a shared interface for action generation and scene dynamics.
PointWorld~\citep{huang2026pointworld} directly predicts metric point
flows conditioned on externally supplied robot motion, rather than
generating that motion from a task instruction.
A 3D-native WAM should place observations, actions, and resulting scene
changes in a shared metric frame, enabling task-conditioned behavior,
consequence prediction, and executable control.

\begin{figure}
    \centering
    \includegraphics[width=\linewidth]{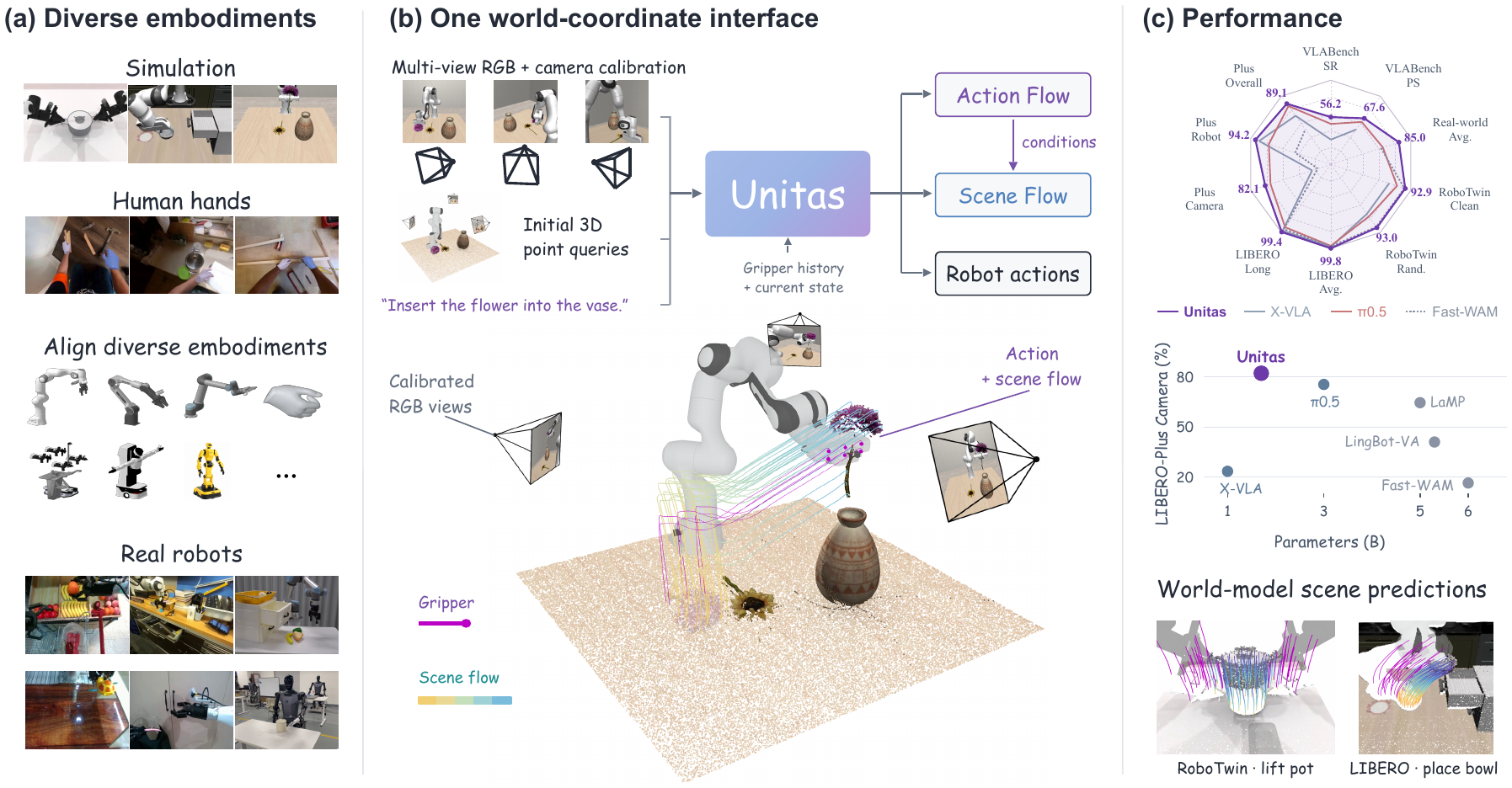}
    \vspace{-0.7cm}
    \caption{\textbf{\method{} is a 3D-native world action model.}
    Observations, action flow, and scene flow share a metric frame within
    each prediction window, with a common representation across embodiments.
    With 1.7B parameters, \method{} achieves leading manipulation results
    among the compared methods and predicts action-conditioned scene
    trajectories.\vspace{-0.4cm}}
    \label{fig:physwam_overview}
\end{figure}

We introduce \method{}, \textbf{UNI}fied \textbf{T}rajectories for
\textbf{A}ction and \textbf{S}cene flow.
To our knowledge, it is the first 3D-native world action model that
aligns observations, actions, and scene dynamics in a shared metric
frame within each prediction window, using a common representation
across robot embodiments and human hands, as illustrated in
Fig.~\ref{fig:physwam_overview}.

Realizing this 3D-native world action model presents three challenges.
First, visual observations must be grounded in metric space even
when depth is unavailable. We propose \emph{world-aligned 3D position embeddings} derived from observed depth or learned weights over
candidate positions along calibrated camera rays. Second, human hands and robot grippers have different morphologies
and control interfaces. We propose \emph{action flow}, which represents their motion as trajectories of corresponding 3D points in the shared frame. Third, trajectories differ in frame rate and duration across data sources. Representing each timestep as a token allocates backbone capacity
across unequal physical-time intervals. We propose a \emph{physical-time trajectory tokenizer}, which aligns each trajectory to elapsed time and encodes the whole trajectory as one token before the backbone.
\emph{Scene flow} describes scene-point displacements decoded from trajectory
tokens conditioned on action flow. 

In the backbone, observation tokens
are grounded in world coordinates, while action-flow tokens across
embodiments and scene-flow tokens are anchored at each point's current
3D position, allowing the network to model motion through features
associated with fixed spatial anchors.
Given the instruction and current observations, a layer-coupled
Mixture-of-Transformers jointly models point trajectories and executable
commands. This supports two modes: \emph{policy mode} directly generates
executable actions, while \emph{simulator mode} predicts scene changes
conditioned on externally supplied action flow.
A causal attention mask enables key--value caching of shared context
across denoising steps, reducing repeated computation and accelerating
inference.


We evaluate \method{} on action-conditioned scene prediction and
robotic manipulation.
In simulator mode, it leads the compared methods across all four
RoboTwin scene-prediction metrics, reducing Moving ADE, Moving FDE,
and Static ADE by 35\%, 21\%, and 49\%, respectively, relative to
PointWorld~\citep{huang2026pointworld}.
In policy mode, it achieves 99.8\% success on LIBERO and outperforms
larger video-based WAMs on RoboTwin.
It leads the compared methods in VLABench success rate and progress
score, and reaches 89.1\% success on LIBERO-Plus with strong robustness
to camera and robot-state changes.
Real-robot success reaches 90\% on \emph{Organize Books} and 65\% on
\emph{Pack the Ball}, exceeding both baselines.
Ablations show that world-aligned visual grounding, action- and scene-flow
supervision, and motion history improve policy performance and robustness.

Our contributions are:\nopagebreak[4]
\begin{itemize}
    \item To our knowledge, \method{} is the first 3D-native world action
    model that aligns observations, actions, and scene dynamics in a
    shared metric frame, using a common representation across robot
    embodiments and human hands.

    \item Action flow represents human hands and robot grippers through
    point trajectories, while scene flow represents the environmental
    response conditioned on that motion. Both policy and simulator modes
    use this interface.

    \item World-aligned positional embeddings ground visual observations
    with or without depth, and a physical-time trajectory tokenizer
    encodes each point trajectory as one token spatially anchored at its
    current 3D position.

    \item With 1.7B parameters, \method{} achieves leading action-conditioned
    scene prediction and strong manipulation performance in simulation
    and on real hardware.
\end{itemize}

\vspace{-0.3cm}

\section{Related Work}
\label{sec:related_work}
\vspace{-0.2cm}

\noindent\textbf{Video world--action modeling.}
Video-based policies use future prediction for control \citep{hu2024vpp},
with early WAMs coupling prediction and action learning through shared
representations and generative objectives
\citep{li2025uva,zhu2025uwm,cen2025worldvla,xu2025imaginative}.
Video-generation priors subsequently support joint video--action learning
and future-conditioned policies
\citep{genieenvisioner,shen2025videovla,bi2025motus,cosmospolicy,li2026lingbotva,wang2026eva},
while heterogeneous robot and video-only data support cross-task and
cross-embodiment transfer \citep{ye2026dreamzero,lingbotva2}.
Efficient variants omit future generation at deployment or replace it
with latent subgoals \citep{yuan2026fastwam,chen2026lawam}.
These models primarily represent world evolution in images or visual
latents. Reported sensitivity to viewpoint and robot initial-state changes
\citep{zhang2026wamgeneralization,hong2026walqr} highlights the importance
of grounding perception, world evolution, and executable actions in a
shared physical space.

\noindent\textbf{Policy learning in 3D space.}
Explicit 3D representations improve spatial understanding for manipulation
\citep{shridhar2023peract,ze2024threediffusion,zhen2024threedvla,qu2025spatialvla}.
Geometry-aware world models augment visual prediction with image-aligned
geometric signals \citep{zhen2025tesseract,guo2026xwam,zhao2026rynnworld}.
In policy learning, LaMP~\citep{wang2026lamp} and
GeomVLA~\citep{xiong2026geomvla} use intermediate motion features from
trajectory denoising to condition action generation. PointWorld~\citep{huang2026pointworld}
follows the complementary direction, predicting scene flow conditioned
on robot motion for planning. \method{} jointly models agent motion and
its conditional scene response through point flows, while supporting
direct command execution without generating scene flows.

\noindent\textbf{Unified action representations for manipulation.}
Cross-embodiment learning must reconcile different morphologies and
control interfaces \citep{chi2024umi}. Generalist policies use
embodiment-specific prompts, canonical state--action formats, and aligned
end-effector conventions
\citep{liu2024rdt,zheng2025xvla,yuan2026qwenrobotmanip,xiaomi2026robotics1}.
Point flows provide transferable motion targets across embodiments
\citep{yuan2024generalflow,bharadhwaj2024track2act,xu2024im2flow2act},
connecting object-motion prediction to control
\citep{zhi2025threedflowaction,dharmarajan2025dream2flow} and representing
robot and object trajectories \citep{huang2026pointworld,lin2026chronoflow}.
\method{} unifies human-hand, robot-gripper, and scene trajectories with
a shared physical-time tokenizer.

\vspace{-0.3cm}
\section{Methodology}
\label{sec:method}
\vspace{-0.2cm}

\begin{figure}[!t]
    \centering
    \includegraphics[width=\linewidth]{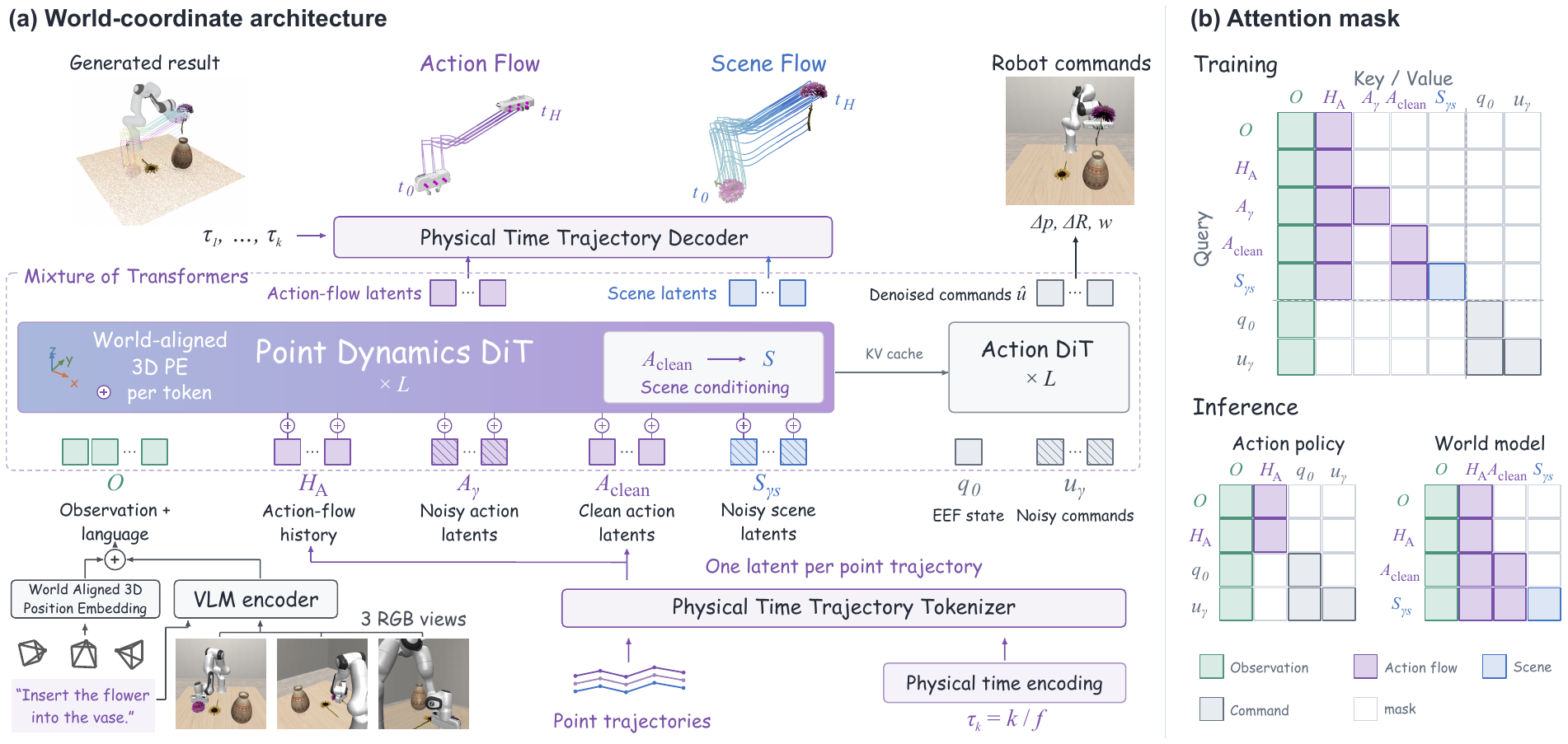}
    \vspace{-0.7cm}
    \caption{\textbf{\method{} architecture and attention masks.}
    Left: The point dynamics expert models both action and scene flows, while the action
    expert generates the end-effector action chunk. Scene flow generation is conditioned
    on clean demonstrated action flow during training and generated or supplied action flow at inference.
    Right: Colored/white cells indicate
    allowed/blocked attention. Policy-only inference omits both future action-flow and scene-flow tokens.}
    \vspace{-0.3cm}
    \label{fig:architecture}
    \label{fig:attention_mask}
\end{figure}

We first revisit video-based world action modeling, then formulate
\method{} in a shared metric 3D frame, as illustrated in
Fig.~\ref{fig:architecture}.

\vspace{-0.2cm}
\subsection{Revisiting Video World Action Models}
\label{sec:method:video_wam}
\vspace{-0.2cm}

Let $o$ denote current image observations, $\ell$ the instruction,
$a_{1:H}$ an action chunk, and $v_{1:T}$ future images.
Video-based WAMs represent future interactions through images or visual
latents. For models that condition actions on predicted futures, the
action distribution can be written as
\begin{equation}
    p(a_{1:H}\mid o,\ell)
    =\int p(v_{1:T}\mid o,\ell)\,
    p(a_{1:H}\mid o,\ell,v_{1:T})\,dv_{1:T}.
    \label{eq:video_wam}
\end{equation}
Future videos and actions may be generated jointly or sequentially.
Fast-WAM~\citep{yuan2026fastwam} instead retains video co-training but
omits future-video generation at inference:
\begin{equation}
    p_{\theta}(a_{1:H}\mid o,\ell)
    =p_{\theta}\!\left(a_{1:H}\mid z(o,\ell)\right),
    \qquad
    \mathcal L
    =\mathcal L_{\mathrm{act}}+\lambda\mathcal L_{\mathrm{vid}}.
    \label{eq:fastwam}
\end{equation}
Its video backbone encodes the current observation and instruction into
$z(o,\ell)$ in one forward pass, while the action expert still performs
iterative denoising. This removes future-video sampling, but the shared
representation remains image-based, without an explicit metric 3D frame.

\vspace{-0.2cm}
\subsection{3D-Native World Action Formulation}
\label{sec:method:formulation}
\vspace{-0.2cm}

At time $t$, context $\mathcal C$ comprises the instruction $\ell$,
multi-view observations $\mathcal O_t=\{\mathcal O_{t,i}\}_{i=1}^{N_v}$,
camera calibration $\{\mathbf K_i,\mathbf T^W_{t,i}\}_{i=1}^{N_v}$,
and historical action flow $\mathcal H_{\mathrm A}$.
Here $\mathbf K_i$ denotes camera intrinsics, and
$\mathbf T^W_{t,i}\in\mathrm{SE}(3)$ maps camera coordinates into the
metric frame $\mathcal F^W$, fixed within each prediction window.
Action prediction additionally uses the current end-effector state
$\mathbf q_0$.

Given query points and future timestamps, we predict action flow
$\mathbf A$ for the interacting body and scene flow $\mathbf S$ for
the environment, both as displacement trajectories in $\mathcal F^W$.
A separate action branch predicts an executable end-effector action
chunk $\mathbf u$. Our model factorizes their joint distribution as
\begin{equation}
    p_{\theta}(\mathbf A,\mathbf u,\mathbf S\mid\mathcal C,\mathbf q_0)
    =
    p_{\theta}(\mathbf A\mid\mathcal C)\,
    p_{\theta}(\mathbf u\mid\mathcal C,\mathbf q_0)\,
    p_{\theta}(\mathbf S\mid\mathbf A,\mathcal C).
    \label{eq:physwam_factorization}
\end{equation}
Action flow and the action chunk can be sampled in parallel; scene prediction
then conditions on generated action flow $\widehat{\mathbf A}$ or an
externally supplied trajectory $\mathbf A_{\mathrm{ext}}$ in simulator
mode. Policy-only execution evaluates only
$p_{\theta}(\mathbf u\mid\mathcal C,\mathbf q_0)$, omitting both point flows.
The following sections describe spatial grounding, the shared trajectory
representation, and the coupled experts that implement these factors.

\vspace{-0.2cm}

\subsection{World-Aligned Observations}
\label{sec:method:observation}
\vspace{-0.2cm}

\noindent\textbf{Frame alignment.}
Within each prediction window, observations, robot or human motion, and
scene trajectories share one fixed metric frame.
Robot demonstrations use a robot-base or shared embodiment frame.
Egocentric human demonstrations use the camera frame of the window anchor,
held fixed throughout the window.
Appendix~\ref{app:dataset_visualization} lists the dataset conventions.

\noindent\textbf{World-aligned 3D position embedding.}
Each visual token receives a position embedding in the world frame.
For a view with valid depth,
$\mathcal O_{t,i}=(\mathbf I_{t,i},\mathbf D_{t,i})$ contains the RGB
image and the depth along the optical axis.
The vision encoder maps the image to a token grid
$\mathbf F_{t,i}=\operatorname{Enc}_{\mathrm{vis}}(\mathbf I_{t,i})$.
We back-project each pixel and transform it into world coordinates,
\begin{equation}
    \mathbf P_{\mathrm{camera},i}(u,v)
    =\mathbf D_{t,i}(u,v)\,\mathbf K_i^{-1}[u,v,1]^\top,
    \qquad
    \begin{bmatrix}
        \mathbf P_{\mathrm{world},i}(u,v)\\ 1
    \end{bmatrix}
    =\mathbf T^W_{t,i}
    \begin{bmatrix}
        \mathbf P_{\mathrm{camera},i}(u,v)\\ 1
    \end{bmatrix}.
    \label{eq:world_coordinate_map}
\end{equation}
Here, $(u,v)$ denotes pixel-center coordinates, and the time index is
omitted from the coordinate maps.
A fixed Fourier map $\boldsymbol\psi$ encodes the world position.
Inside each patch, $\operatorname{Enc}_{\mathrm{3D}}$ applies
$\boldsymbol\psi$ and projects the result to the visual-token dimension:
\begin{equation}
    \mathbf E_{\mathrm{pos},t,i}
    =\operatorname{Enc}_{\mathrm{3D}}(\mathbf P_{\mathrm{world},i}),
    \qquad
    \widetilde{\mathbf F}_{t,i}
    =\mathbf F_{t,i}+\mathbf E_{\mathrm{pos},t,i}.
    \label{eq:depth_position_embedding}
\end{equation}

\noindent\textbf{Spatial grounding without observed depth.}
Without depth, the position embedding is computed from a weighted
combination of positional features along the camera ray.
For a patch centered at $(u,v)$, we draw $M=64$ depths uniformly from
$0.1$ to $2.0\,\mathrm{m}$, back-project them by
Eq.~\ref{eq:world_coordinate_map}, and score them with the patch feature
$\mathbf f$:
\begin{equation}
\begin{aligned}
    \alpha_m
    &=\operatorname{softmax}_{m}
    \left(
        \frac{(\mathbf W_q\mathbf f)^\top
        \mathbf W_k\boldsymbol\psi(\mathbf p_{\mathrm{camera},m})}
        {\sqrt{d}}
    \right),\\
    \mathbf E_{\mathrm{ground}}
    &=\mathbf W_o\sum_{m=1}^{M}
        \alpha_m\mathbf W_v
        \boldsymbol\psi(\mathbf p_{\mathrm{world},m}).
\end{aligned}
\label{eq:frustum_routing}
\end{equation}
Here $\mathbf f\in\mathbb R^{d_f}$ is the patch feature and
$\boldsymbol\psi(\cdot)\in\mathbb R^{d_\psi}$, with
$\mathbf W_q\in\mathbb R^{d\times d_f}$,
$\mathbf W_k\in\mathbb R^{d\times d_\psi}$, and
$\mathbf W_v\in\mathbb R^{d\times d_\psi}$.
Keys encode camera-frame positions, and values encode the same candidates
in the world frame.
$\mathbf E_{\mathrm{ground}}$ is added to the visual tokens in place of
$\mathbf E_{\mathrm{pos},t,i}$ in Eq.~\ref{eq:depth_position_embedding}.
Where depth is reliable, we use depth-derived embeddings and supervise
the routing weights by matching Fourier features:
\begin{equation}
    \mathcal L_{\mathrm{PE}}
    =\sum_{i=1}^{N_v}
    \mathbb E_{\mathcal R\in\mathcal V_i}
    \left\|
        \sum_{m=1}^{M}\alpha_m
        \boldsymbol\psi(\mathbf p_{\mathrm{world},m})
        -\operatorname{sg}\!\left(
            \operatorname{Avg}_{(u,v)\in\mathcal R}
            \boldsymbol\psi(\mathbf P_{\mathrm{world},i}(u,v))
        \right)
    \right\|_2^2 .
    \label{eq:position_distillation}
\end{equation}
Here $\mathcal V_i$ contains regions with reliable depth in view $i$,
and $\operatorname{sg}$ denotes stop-gradient. Only valid-depth pixels
contribute to the target average.
With this module, \method{} pretrains jointly on RGB and RGB-D data, and depth may be absent at inference.
Appendix~\ref{app:missing_depth} evaluates the router when depth is
removed at test time.

\vspace{-0.2cm}
\subsection{Physical-Time Point-Flow Representation}
\label{sec:method:flow_representation}
\vspace{-0.2cm}

\noindent\textbf{Action flow and scene flow.}
In the same world frame, action and scene flows describe displacement
from each query point's current position.
For query $n$ at $\mathbf x_{t,n}$, the target is
$\{\Delta\mathbf x_{n,k}\}_{k=1}^{H}$, with
$\Delta\mathbf x_{n,k}=\mathbf x_{t+\tau_k,n}-\mathbf x_{t,n}$.
The elapsed time is $\tau_k$, and point identity is preserved throughout
the trajectory.

For robots, we sample $N_g=20$ points from the URDF-defined gripper
geometry of each arm and retain their link-local coordinates.
Forward kinematics maps these points into the world frame at each
recorded joint configuration, preserving correspondence through arm and
finger motion.
Human demonstrations use 21 MANO-derived keypoints~\citep{romero2017mano}
per observed hand. Following PointWorld~\citep{huang2026pointworld},
we mask robot pixels using the URDF and current joint state, back-project
the remaining RGB-D pixels, and sample $N_s$ scene query points. Simulator states supply ground-truth correspondences
where available and scene-flow loss is applied only to valid trajectories.
At inference, scene prediction requires current 3D query coordinates
from depth or supplied geometry.

\noindent\textbf{Physical-time trajectory tokenizer.}
To better align different trajectories within diverse datasets, we encode time in seconds rather than frame indices, using $\tau_k=k/f$
at sampling rate $f$.
A fixed basis $\boldsymbol\phi_{\mathrm{time}}$ of $\tau/T_{\max}$, with
$T_{\max}=3\,\mathrm{s}$ shared across datasets, gives equal elapsed times
the same features; history uses negative offsets.
An input MLP maps each displacement to $\mathbf h_{n,k}$, and temporal
MLPs map $\boldsymbol\phi_{\mathrm{time}}(\tau_k)$ to FiLM parameters
$(\mathbf a_k,\mathbf b_k)$ for the encoder and the decoder.
The encoder pools the trajectory into one latent, and the decoder reads
that latent back at a requested time:
\begin{equation}
\begin{aligned}
    \mathbf z_n
    &=\frac{1}{\sqrt H}\sum_{k=1}^{H}
    \left[(\mathbf 1+\mathbf a^{\mathrm{enc}}_k)\odot
    \mathbf h_{n,k}+\mathbf b^{\mathrm{enc}}_k\right],\\
    \widehat{\Delta\mathbf x}_{n,k}
    &=\operatorname{MLP}_{\mathrm{out}}
    \left[(\mathbf 1+\mathbf a^{\mathrm{dec}}_k)\odot
    \mathbf z_n+\mathbf b^{\mathrm{dec}}_k\right].
\end{aligned}
\label{eq:physical_time_flowchunk}
\end{equation}
Padding is excluded from the sum.
Action flow, scene flow, and history share this tokenizer, so each point
is one token regardless of trajectory length or sampling density.
Flow tokens are anchored only at current world-frame points:
positional embeddings as in Eq.~\ref{eq:depth_position_embedding}
are added throughout denoising. Decoding gives
$\widehat{\mathbf x}_{t+\tau_k,n}
=\mathbf x_{t,n}+\widehat{\Delta\mathbf x}_{n,k}$.
Appendix~\ref{app:trajectory_tokenizer} gives the basis and the
reconstruction stage.

\begin{figure}[!t]
    \centering
    \includegraphics[width=\linewidth]{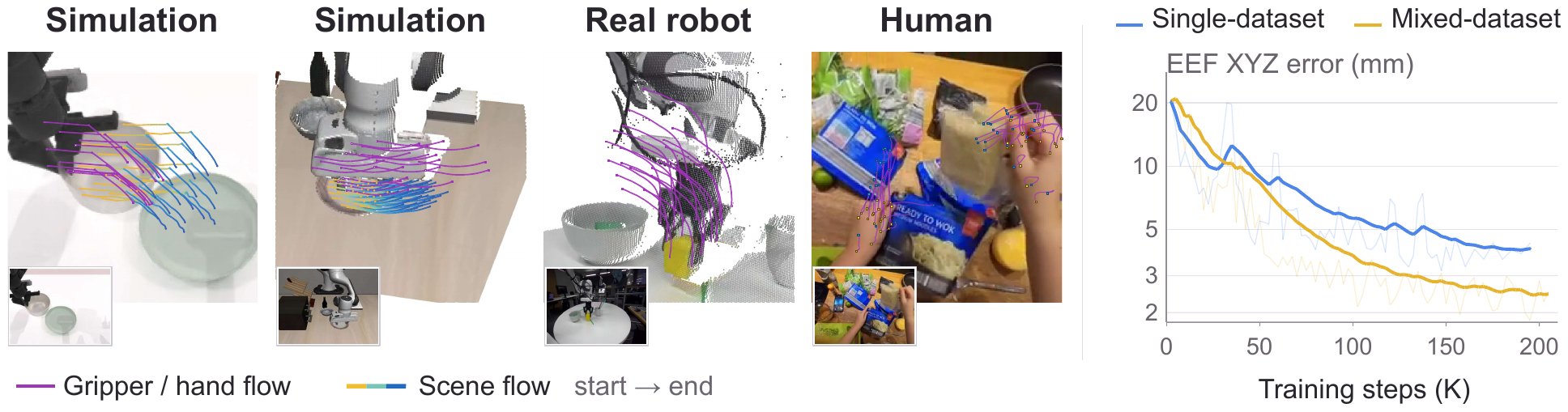}
    \vspace{-0.8cm}
    \caption{\textbf{Shared point-flow representations across data sources.}
    Left: gripper/hand trajectories (purple) and scene trajectories
    (yellow--blue), with RGB insets. Right: RoboTwin end-effector XYZ
    error (mm). Thick curves show exponential moving averages (EMA)
    of the thin curves. \vspace{-0.2cm}}
    \label{fig:experiments:mixed_dataset_training}
\end{figure}

\vspace{-0.2cm}

\subsection{Model Architecture}
\label{sec:method:architecture}
\vspace{-0.2cm}

\noindent\textbf{Coupled experts.}
\method{} uses a Mixture-of-Transformers~\citep{liang2024mixture}
with two diffusion transformers~\citep{peebles2023scalable}:
a point dynamics expert for action and scene flows, and an action expert for the
end-effector action chunk $\mathbf u$. The point dynamics expert processes spatially grounded
primary-view VLM tokens, auxiliary-view features from the shared vision
tower, and historical and future flow latents. At each layer, the smaller
action expert reads the dynamics expert's observation/language keys and
values (KV), together with its own current-state and noisy command tokens.
It predicts relative end-effector translation, rotation, and gripper
openness at each future timestep.

\noindent\textbf{Attention structure.}
Let $O$ and $H_{\mathrm A}$ be the observation/language and history
tokens, and let $A_\gamma$, $A_{\mathrm{clean}}$, and $S_{\gamma_{\mathrm s}}$
be the noisy action-flow, clean action-flow, and noisy scene latents.
The context $[O,H_{\mathrm A}]$ is bidirectional.
Each action-flow group attends to this context and to itself.
Scene tokens attend to the context, to $A_{\mathrm{clean}}$, and to one
another, but not to $A_\gamma$, so scene prediction conditions on action
flow without entering action-flow denoising.
Action chunk tokens attend to the observation KV, the current end-effector
state, and one another.
They do not attend to future action-flow or scene-flow tokens, so
policy-only inference can skip both future-flow branches, as shown in
Fig.~\ref{fig:architecture}.

\noindent\textbf{Training.}
We first train the trajectory tokenizer to reconstruct historical and
future action flow and available scene trajectories, then freeze it.
Both experts learn through rectified flow~\citep{liu2022rectifiedflow}.
For a clean target $\mathbf y$, either a normalized trajectory latent or
an action chunk, and noise
$\boldsymbol\epsilon\sim\mathcal N(\mathbf 0,\mathbf I)$, we form
$\mathbf y_\gamma=(1-\gamma)\boldsymbol\epsilon+\gamma\mathbf y$ and
regress $\mathbf y-\boldsymbol\epsilon$.
Action flow and action chunk share $\gamma$ but use independent noise, while scene
flow draws $\gamma_{\mathrm s}$ independently.
During training, scene prediction conditions on clean action-flow latents
from demonstrations. The objective is
\begin{equation}
    \mathcal L
    =\lambda_{\mathrm A}\mathcal L_{\mathrm A}
    +\lambda_{\mathrm s}\mathcal L_{\mathrm{scene}}
    +\lambda_{\mathrm{ctrl}}\mathcal L_{\mathrm{ctrl}}
    +\lambda_{\mathrm{PE}}\mathcal L_{\mathrm{PE}}.
    \label{eq:training_objective}
\end{equation}
The first three terms use mean squared error (MSE) losses to supervise
action flow, scene flow, and the action chunk, with
$\lambda_{\mathrm A}=\lambda_{\mathrm s}=\lambda_{\mathrm{ctrl}}=1$.
$\mathcal L_{\mathrm{PE}}$ is Eq.~\ref{eq:position_distillation}, with
$\lambda_{\mathrm{PE}}=0.1$.
Each loss uses only available annotations, with missing targets and
absent arm slots masked.

\noindent\textbf{Inference.}
Action flow and action chunk are denoised in parallel on one Euler schedule.
Generated action flow, or an externally supplied trajectory in simulator
mode, then conditions scene denoising.
The context and clean action-flow keys and values are cached and reused
across scene-denoising steps.
Policy-only execution keeps the observation and history context and
denoises only $\mathbf u$, skipping both point dynamics branches.
Appendix~\ref{app:implementation} gives the schedule, widths, and
optimization.

\vspace{-0.3cm}

\section{Policy Evaluation}
\label{sec:experiments}
\label{sec:experiments:policy}
\vspace{-0.2cm}

Our policy evaluation tests whether representing interaction in a shared
metric space supports effective robot control. We examine task completion
across simulation benchmarks, robustness when observations and initial
states change, and execution on real-world robots.

\vspace{-0.2cm}

\subsection{Experimental Setup}
\label{sec:experiments:setup}
\vspace{-0.2cm}

\noindent\textbf{Training settings.}
We initialize the vision-language backbone from
Florence-2~\citep{xiao2023florence2} and randomly initialize all remaining
components. We consider two training settings: (1) \emph{with pretraining},
\method{} is pretrained on a mixture of real-robot, simulation, and
human hand demonstrations, then fine-tuned on each downstream benchmark
or dataset; (2) \emph{without pretraining}, \method{} (w/o Pretrain) is
fine-tuned directly on each target benchmark or dataset without the
mixed-dataset pretraining stage. More detailed training configurations
are provided in the appendix.

\noindent\textbf{Baselines.}
We compare against VLA policies
\citep{zheng2025xvla,sun2026vlajepa,black2024pizero,physicalintelligence2025pi05,abotm0},
video-based WAMs \citep{cosmospolicy,genieenvisioner,bi2025motus,li2026lingbotva,yuan2026fastwam}, and
3D-aware policies
\citep{qu2025spatialvla,wang2026lamp,guo2026xwam,zhou2026dexbev}.

\vspace{-0.2cm}

\subsection{Simulation Benchmarks}
\label{sec:experiments:simulation}
\vspace{-0.2cm}

\begin{table}[!b]
    \centering
    \small
    \setlength{\tabcolsep}{2.5pt}
    \renewcommand{\arraystretch}{0.93}
    \vspace{-0.5cm}
    \caption{Policy performance (\%, $\uparrow$). Our 1.7B model is evaluated
    with and without mixed-dataset pretraining. Bold and underlined scores
    denote the best and second-best results, respectively.
    LIBERO-Plus ranks use one-decimal precision.
    Dashes denote unavailable results.}
    \label{tab:experiments:simulation}
    \begin{adjustbox}{max width=0.94\linewidth}
    \begin{tabular}{@{}l*{12}{c}@{}}
        \toprule
        \multirow{2}{*}{Method}
        & \multirow{2}{*}{\shortstack{Params\\(B)}}
        & \multicolumn{5}{c}{LIBERO} & LIBERO-Plus
        & \multicolumn{2}{c}{RoboTwin 2.0} & \multicolumn{3}{c}{VLABench} \\
        \cmidrule(lr){3-7}\cmidrule(lr){9-10}\cmidrule(lr){11-13}
        & & Spatial & Object & Goal & 10 & Avg. & Overall & Clean & Rand. & SR & PS & IS \\
        \midrule
        \multicolumn{13}{l}{\textit{VLA Policies}} \\
        X-VLA & 1.0 & 98.2 & 98.6 & 97.8 & 97.6 & 98.1 & 71.4 & 72.88 & 72.84 & 29.4 & 51.2 & \underline{70.2} \\
        VLA-JEPA & 2.0 & 96.2 & 99.6 & 97.2 & 95.8 & 97.2 & 77.9 & -- & -- & -- & -- & -- \\
        $\pi_0$ & 3.0 & 96.8 & 98.8 & 95.8 & 85.2 & 94.15 & 53.6 & 65.92 & 58.40 & 29.4 & 44.1 & 55.0 \\
        $\pi_{0.5}$ & 3.0 & 98.8 & 98.2 & 98.0 & 92.4 & 96.9 & 85.7 & 82.74 & 76.76 & 48.1 & 62.3 & 64.9 \\
        ABot-M0 & 4.0 & 98.8 & \underline{99.8} & 99.0 & 96.6 & 98.6 & 80.5 & 80.42 & 81.16 & -- & -- & -- \\
        \midrule
        \multicolumn{13}{l}{\textit{Video-Based WAMs}} \\
        Cosmos Policy & 2.0 & 98.1 & \textbf{100.0} & 98.2 & 97.6 & 98.5 & 82.2 & -- & -- & -- & -- & -- \\
        GE-Act & 2.2 & 98.2 & 97.6 & 95.8 & 94.4 & 96.5 & 80.3 & -- & -- & -- & -- & -- \\
        LingBot-VA & 5.3 & 98.5 & 99.6 & 97.2 & 98.5 & 98.5 & 69.5 & \underline{92.93} & 91.55 & -- & -- & -- \\
        Fast-WAM & 6.0 & 98.2 & \textbf{100.0} & 97.0 & 95.2 & 97.6 & 51.5 & 91.88 & \underline{91.78} & -- & -- & -- \\
        Motus & 8.0 & 96.8 & \underline{99.8} & 96.6 & 97.6 & 97.7 & -- & 88.66 & 87.02 & -- & -- & -- \\
        \midrule
        \multicolumn{13}{l}{\textit{3D-Aware Policies}} \\
        SpatialVLA & 3.5 & 88.2 & 89.9 & 78.6 & 55.5 & 78.1 & -- & -- & -- & -- & -- & -- \\
        LaMP & 5.0 & \underline{99.4} & \underline{99.8} & 97.4 & 96.7 & 98.3 & 79.3 & -- & -- & -- & -- & -- \\
        X-WAM & 6.7 & -- & -- & -- & -- & -- & -- & 89.8 & 90.7 & -- & -- & -- \\
        Dex-BEV & -- & 98.2 & 98.0 & 97.8 & 97.0 & 97.8 & -- & 76.0 & 42.0 & -- & -- & -- \\
        \midrule
        \method{} (w/o Pretrain) & 1.7 & \textbf{99.8} & \underline{99.8} & \underline{99.4} & \underline{99.0} & \underline{99.5} & \underline{86.5} & 91.08 & 90.74 & \underline{50.3} & \underline{64.3} & \textbf{71.2} \\
        \method{} & 1.7 & \textbf{99.8} & \textbf{100.0} & \textbf{100.0} & \textbf{99.4} & \textbf{99.8} & \textbf{89.1} & \textbf{92.94} & \textbf{93.04} & \textbf{56.2} & \textbf{67.6} & 69.3 \\
        \bottomrule
    \end{tabular}
    \end{adjustbox}
    \vspace{-0.15cm}
\end{table}

We evaluate \method{} on single-arm and bimanual manipulation benchmarks.
LIBERO~\citep{liu2023libero} includes Spatial, Object, Goal, and
long-horizon LIBERO-10 suites; RoboTwin 2.0~\citep{chen2025robotwin}
covers 50 bimanual tasks under Clean and Randomized conditions.
VLABench~\citep{zhang2024vlabench} tests object and instruction
generalization across five tracks, measured by success rate (SR),
progress score (PS), and intention score (IS).
LIBERO-Plus~\citep{fei2025liberoplus} evaluates robustness to seven
types of observation, instruction, and initial-state perturbations.
Detailed protocols and per-task results are provided in the appendix.

\noindent\textbf{Manipulation performance.}
Tab.~\ref{tab:experiments:simulation} reports policy performance across
the evaluated simulation benchmarks. On LIBERO, \method{} achieves a
state-of-the-art average success rate of 99.8\%, with strong results
on spatial and long-horizon tasks. Without mixed-dataset pretraining,
it achieves 99.5\%, also exceeding all baselines.
On RoboTwin 2.0, \method{} achieves 92.94\% and 93.04\% under Clean
and Randomized conditions, respectively. Its 1.7B model outperforms
the compared methods while using fewer parameters than Fast-WAM (6B)
and Motus (8B). On VLABench, it achieves 56.2\% SR compared with
48.1\% for $\pi_{0.5}$, alongside a higher progress score.

\noindent\textbf{Robustness to distribution shifts.}
As shown in Tab.~\ref{tab:experiments:simulation}, \method{}
achieves the highest overall success rate of 89.1\%
on LIBERO-Plus. Appendix~\ref{app:libero_plus} reports the perturbation-wise results.
Under camera perturbations, it achieves 82.1\%, compared with 75.8\%
for Cosmos Policy and 64.5\% for LaMP. Both training settings show
strong viewpoint robustness, consistent with world-coordinate grounding
of visual observations.

Mixed-dataset pretraining improves overall success by 2.6 percentage points, with the largest gains under robot initial-state changes from 84.8\% to 94.2\%, and language perturbations from 77.6\% to 85.0\%. These improvements suggest better generalization to changes in both physical configuration and instruction phrasing.
As illustrated in Fig.~\ref{fig:experiments:mixed_dataset_training},
adding other data sources while keeping the number of RoboTwin samples per batch unchanged yields lower RoboTwin test XYZ error.
Together, these results support positive cross-dataset transfer
under our unified modeling formulation.

\vspace{-0.2cm}

\subsection{Real-world Evaluation}
\label{sec:experiments:real_world}
\vspace{-0.2cm}

We evaluate two real-world tasks: \emph{Organize Books} requires aligning
books with shelf compartments, while \emph{Pack the Ball} requires placing
a ball inside a bag and closing its zipper. We compare against $\pi_{0.5}$
and Fast-WAM using 20 trials per method per task.
As shown in Tab.~\ref{tab:experiments:real_world}, \method{} achieves
90\% and 80\% success on the two tasks, respectively, outperforming
both baselines.
Fig.~\ref{fig:experiments:real_world} shows representative executions.

\begin{table}[!t]
    \centering
    \newsavebox{\realworldimagebox}
    \sbox{\realworldimagebox}{\includegraphics[width=0.65\linewidth]{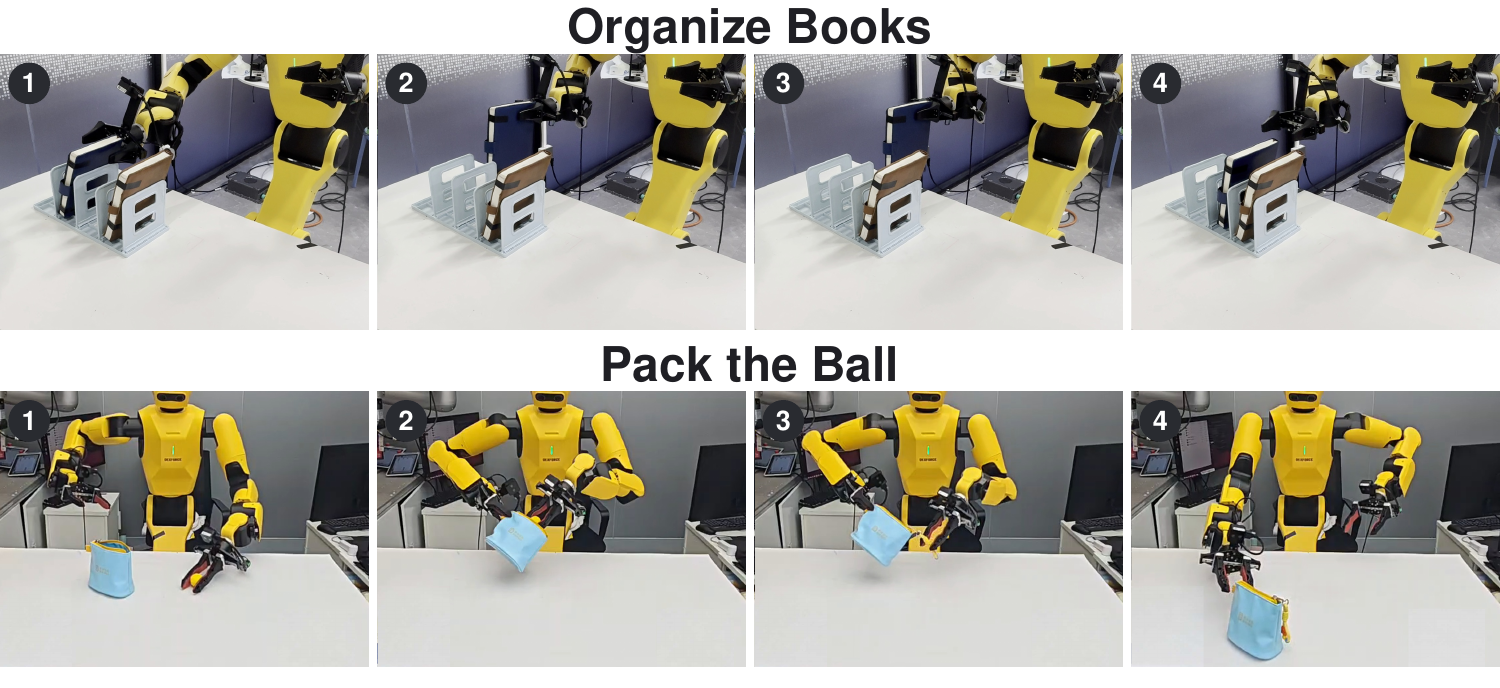}}
    \begin{minipage}[t]{0.65\linewidth}
        \vspace{0pt}
        \usebox{\realworldimagebox}
        \makeatletter\def\@captype{figure}\makeatother
        \vspace{-0.9cm}
        \caption{Qualitative real-robot rollouts on \emph{Organize Books}
        and \emph{Pack the Ball}. Numbered keyframes show task progression
        through precise placement and sequential manipulation.}
        \label{fig:experiments:real_world}
    \end{minipage}\hfill
    \begin{minipage}[t]{0.33\linewidth}
        \vspace{0pt}
        \centering
        \makeatletter\def\@captype{table}\makeatother
        \setlength{\abovecaptionskip}{0pt}
        \caption{\raggedright Quantitative comparison results of real-robot experiments (20 trials per task).}
        \label{tab:experiments:real_world}
        \footnotesize
        \setlength{\aboverulesep}{0pt}
        \setlength{\belowrulesep}{0pt}
        \renewcommand{\arraystretch}{1.0}
        \begin{tabular*}{\linewidth}{@{}l@{\extracolsep{\fill}}r@{}}
            \toprule
            \multicolumn{2}{@{}c@{}}{\colorbox[gray]{0.93}{\makebox[\dimexpr\linewidth-2\fboxsep\relax]{\strut\textbf{Organize Books}}}} \\
            \midrule
            $\pi_{0.5}$ & 15/20 (75\%) \\
            Fast-WAM & 13/20 (65\%) \\
            \method{} & \textbf{18/20 (90\%)} \\
            \midrule
            \multicolumn{2}{@{}c@{}}{\colorbox[gray]{0.93}{\makebox[\dimexpr\linewidth-2\fboxsep\relax]{\strut\textbf{Pack the Ball}}}} \\
            \midrule
            $\pi_{0.5}$ & 11/20 (55\%) \\
            Fast-WAM & 12/20 (60\%) \\
            \method{} & \textbf{16/20 (80\%)} \\
            \bottomrule
        \end{tabular*}
    \end{minipage}
    \vspace{-0.3cm}
\end{table}

\begin{table}[!b]
    \centering
    \small
    \setlength{\tabcolsep}{3pt}
    \renewcommand{\arraystretch}{0.96}
    \vspace{-0.85cm}

    \caption{LIBERO-Plus ablations without mixed-dataset pretraining.
    Checkmarks indicate enabled components.
    Success rates (\%) are reported for each perturbation;
    Overall pools all 10,030 evaluation episodes.}
    \label{tab:experiments:libero_plus_ablation}
    \begin{adjustbox}{max width=0.97\linewidth}
    \begin{tabular}{@{}cccc ccccccc c@{}}
        \toprule
        \multicolumn{4}{c}{Components}
        & \multicolumn{7}{c}{Perturbation Type}
        & Overall \\
        \cmidrule(lr){1-4}
        \cmidrule(lr){5-11}
        \shortstack{Visual\\3D PE}
        & \shortstack{Action-flow\\Loss}
        & \shortstack{History\\Tokens}
        & \shortstack{Scene-flow\\Loss}
        & Camera & Robot & Language & Light
        & Background & Noise & Layout & \\
        \midrule
        & & &
        & 50.47 & 59.48 & 67.14
        & 81.00 & 74.44 & 74.70
        & 71.80 & 67.58 \\

        & $\checkmark$ & $\checkmark$ &
        & 66.73 & 79.29 & 75.21
        & 94.57 & \underline{92.19} & 84.13
        & 84.46 & 81.35 \\

        $\checkmark$ & & $\checkmark$ &
        & 75.05 & 76.13 & 66.17
        & 90.72 & 87.45 & 86.45
        & 75.74 & 78.89 \\

        $\checkmark$ & $\checkmark$ & &
        & 69.04 & 64.71 & 72.41
        & 88.88 & 80.48 & 81.01
        & 79.80 & 75.92 \\

        $\checkmark$ & $\checkmark$ & $\checkmark$ &
        & \underline{79.11} & \underline{84.58} & \underline{77.10}
        & \underline{96.32} & \textbf{94.42} & \underline{89.07}
        & \underline{85.05} & \underline{85.74} \\

        $\checkmark$ & $\checkmark$ & $\checkmark$ & $\checkmark$
        & \textbf{81.86} & \textbf{84.84} & \textbf{77.62}
        & \textbf{96.67} & \textbf{94.42} & \textbf{89.82}
        & \textbf{85.11} & \textbf{86.47} \\
        \bottomrule
    \end{tabular}
    \end{adjustbox}
\end{table}

\vspace{-0.2cm}

\subsection{Ablation Study}
\label{sec:experiments:ablations}
\vspace{-0.2cm}

We ablate visual 3D positional encoding, action-flow supervision,
action-flow history tokens, and scene-flow supervision on LIBERO-Plus without
mixed-dataset pretraining.
As shown in Tab.~\ref{tab:experiments:libero_plus_ablation},
the first three ablations use a reference model without scene-flow
supervision, which achieves 85.74\% overall success.
Removing 3D positional encoding reduces success to 81.35\%,
with the largest drop under camera perturbations, from 79.11\%
to 66.73\%. This highlights the role of world-coordinate grounding
in relating visual observations to robot motion and improving
generalization across camera viewpoints.

Removing action-flow supervision reduces overall success to 78.89\%,
supporting the value of geometric motion supervision for action
learning.
Removing action-flow history tokens reduces overall success to 75.92\%,
with a drop of 19.87 percentage points under robot initial-state
perturbations, suggesting that recent motion provides useful context
beyond the current state.
Removing all three components yields a VLA-style baseline with
67.58\% success. Adding scene-flow supervision to the reference model improves overall
success from 85.74\% to 86.47\%, suggesting that future scene prediction
helps learn representations beneficial to action prediction, consistent
with findings in other WAMs~\citep{yuan2026fastwam}.

\vspace{-0.3cm}

\section{Action-Conditioned World Modeling}
\label{sec:experiments:world_model}
\vspace{-0.2cm}

\noindent\textbf{Evaluation setup.}
We evaluate action-conditioned scene prediction on held-out RoboTwin
rollouts using matched initial observations and robot-motion conditions,
with simulator states providing ground-truth point trajectories.
We compare against PointWorld~\citep{huang2026pointworld}, which directly
predicts 3D trajectories, and video-based Ctrl-World~\citep{ctrlworld}
and X-WAM~\citep{guo2026xwam}. Video predictions are lifted to 3D using
SpatialTrackerV2~\citep{xiao2025spatialtrackerv2}, with
Depth Anything 3~\citep{lin2025depthanything3} for Ctrl-World and native
depth predictions for X-WAM. Both are calibrated using only initial-frame
ground-truth depth.
We report Moving ADE and FDE for trajectory and endpoint accuracy,
Static ADE for background drift, and Moving Acc@1cm for the fraction
of moving-point predictions within 1\,cm of ground truth.
Detailed evaluation protocols are provided in Appendix~\ref{app:matched_action}.

\begin{figure}[!t]
    \centering
    \includegraphics[width=0.9\linewidth]{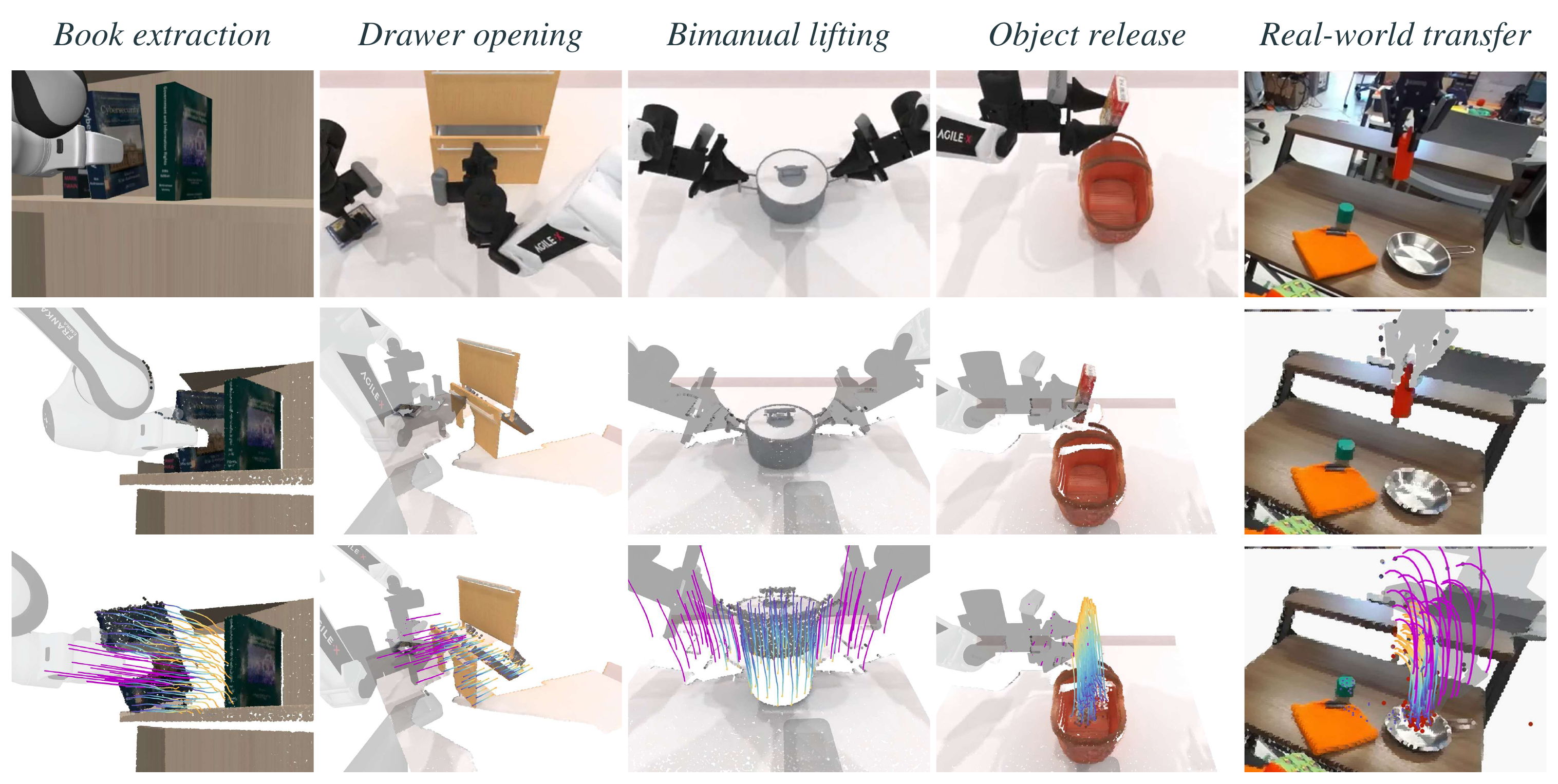}
    \vspace{-0.5cm}

    \caption{\textbf{Action-conditioned scene-flow predictions.}
    Top: observed RGB images. Middle: initial point clouds.
    Bottom: predicted endpoint point clouds and scene trajectories
    (color gradient), conditioned on recorded gripper trajectories
    (purple). Translucent robot meshes indicate the corresponding poses.}
    \label{fig:experiments:scene_flow_qualitative}
    \vspace{-0.5cm}
\end{figure}

\begin{table}[!htbp]
    \centering
    \small
    \setlength{\tabcolsep}{4pt}
    \renewcommand{\arraystretch}{1.00}
    \vspace{-0.5cm}
    \caption{Action-conditioned scene prediction on RoboTwin.
    Errors are in cm ($\downarrow$); Acc@1cm is in \% ($\uparrow$).
    DA3 and STv2 denote Depth Anything 3 and SpatialTrackerV2, respectively.}
    \label{tab:experiments:world_model}
    \begin{adjustbox}{max width=0.97\linewidth}
    \begin{tabular}{@{}lcccc@{}}
        \toprule
        Method & Moving ADE $\downarrow$
        & Static ADE $\downarrow$
        & Moving FDE $\downarrow$
        & Moving Acc@1cm $\uparrow$ \\
        \midrule
        PointWorld & \underline{0.656} & \underline{0.084}
        & \underline{1.050} & \underline{88.22} \\
        Ctrl-World + DA3 + STv2
        & 3.310 & 0.369 & 5.503 & 41.54 \\
        X-WAM + STv2
        & 2.286 & 0.122 & 3.886 & 49.69 \\
        \textbf{\method{}} & \textbf{0.427} & \textbf{0.043}
        & \textbf{0.825} & \textbf{92.31} \\
        \bottomrule
    \end{tabular}
    \end{adjustbox}
    \vspace{-0.3cm}
\end{table}

\noindent\textbf{Action-conditioned scene prediction.}
\method{} achieves the best results across all four metrics, as shown
in Tab.~\ref{tab:experiments:world_model}, reducing Moving ADE from
0.656 to 0.427\,cm relative to PointWorld while improving moving-point
accuracy and reducing static-region drift.
Fig.~\ref{fig:experiments:scene_flow_qualitative} illustrates object
motion and resulting spatial configurations conditioned on gripper
trajectories. The simulation examples use training clips and DROID
example shows real-world scene prediction without real-world
scene-flow supervision.

\noindent\textbf{Action-scene consistency.}
We test whether predicted scene changes agree with the consequences of
the model's own actions on three RoboTwin tasks. \method{} achieves the
best consistency between predicted scene evolution and executed actions
among the compared WAMs. Appendix~\ref{app:future_action_alignment}
provides the evaluation details and full results.

\vspace{-0.3cm}
\section{Conclusion}
\label{sec:conclusion}
\vspace{-0.2cm}

\method{} advances a physical-space formulation of world action modeling:
what the robot observes, how it moves, and how the scene responds are expressed
within a common metric reference. Making interaction geometry an explicit
prediction target connects world modeling with action learning and provides
a shared representation across differing embodiments and observation views.
The strong policy performance observed in simulation supports the value of
this formulation for manipulation. More broadly, this work positions
physical interaction as a unifying basis for embodied learning, with robot
behavior and environmental change represented as complementary parts of the
same process. Extending this foundation to reliable prediction over longer
interactions and broader real-world deployment is a promising direction
for general-purpose robot intelligence.

\subsection*{AI use statement}
Generative AI tools were used to assist with literature review to identify related work and with language polishing.

\bibliography{references}
\bibliographystyle{iclr2027_conference}

\clearpage
\appendix
\providecommand{\method}{Unitas}
\raggedbottom

\section{Implementation and Training Details}
\label{app:experiments}
\label{app:experiments:data}


\subsection{Data Mixture and Temporal Sampling}
\label{app:temporal_sampling}
\label{app:dataset_visualization}

\begin{figure}[!htbp]
    \centering
    \includegraphics[width=\linewidth]{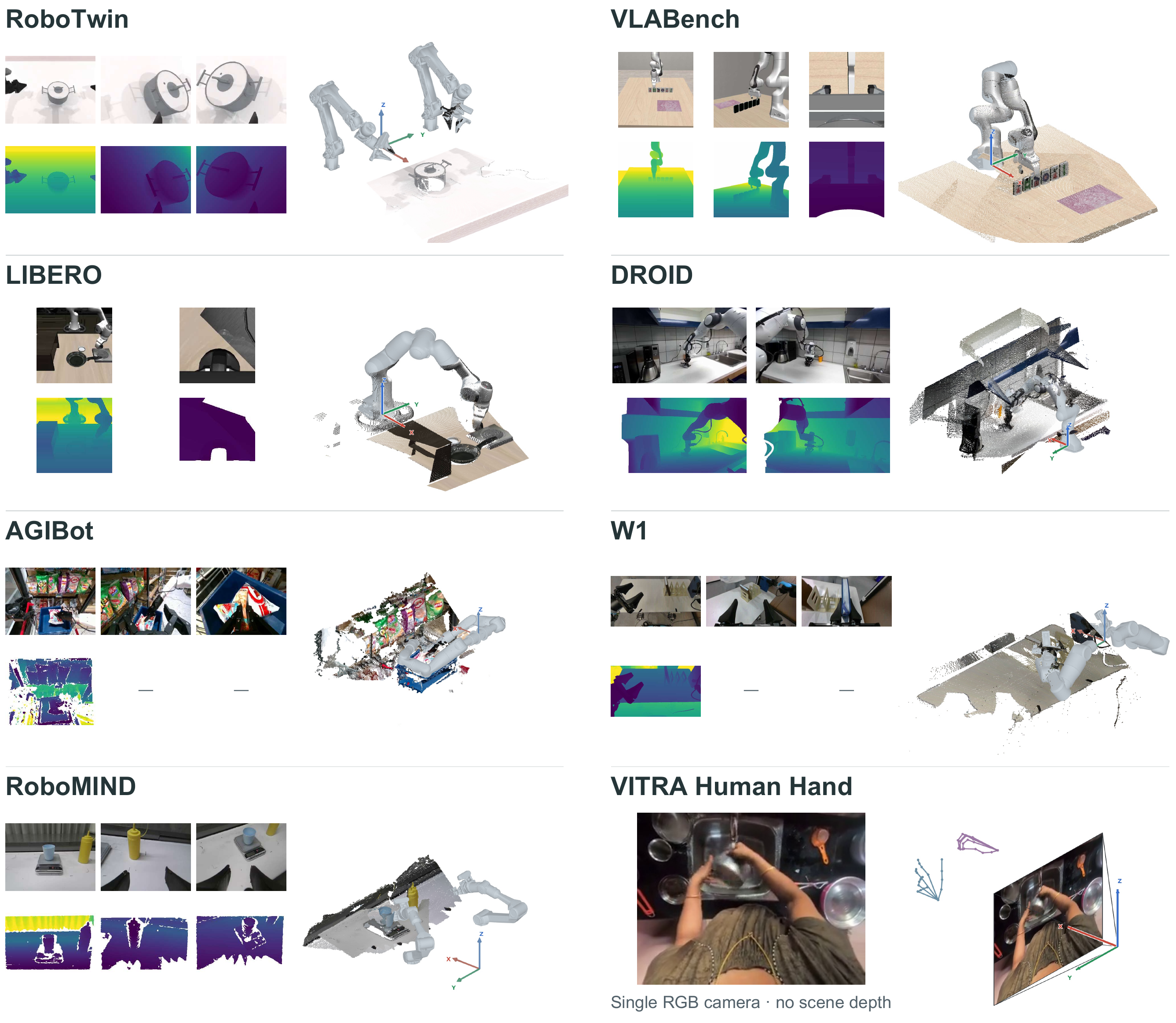}
    \caption{\textbf{Representative observations and aligned geometry from
    pretraining data.} Robot panels show RGB, depth where available, and
    scene points with robot overlays. VITRA shows egocentric RGB and hand
    landmarks.}
    \label{fig:app:dataset_visualization}
\end{figure}

\paragraph{Heterogeneous data mixture.}
We pretrain on a mixture of nine data sources:
RoboTwin~\citep{chen2025robotwin}, LIBERO~\citep{liu2023libero},
VLABench~\citep{zhang2024vlabench}, DROID~\citep{khazatsky2024droid},
AgiBot~\citep{bu2025agibot_arxiv}, RoboMIND~\citep{wu2024robomind},
RH20T~\citep{fang2023rh20t}, VITRA~\citep{li2025vitra}, and our
self-collected W1 data. Fig.~\ref{fig:app:dataset_visualization} shows
representative observations and aligned geometry.

\paragraph{Dataset-specific coordinate conventions.}
Robot data use a base or shared embodiment frame. Where specified by the
bimanual adapter, the origin is the arm-base midpoint, with orientation
inherited from the shared root or torso. W1 training coordinates are
centered at the median end-effector trajectory midpoint. For VITRA,
source camera poses transform hand positions into the current anchor's
camera frame, held fixed throughout the window, with
$(x_M,y_M,z_M)=(z_C,-x_C,-y_C)$.

\paragraph{Temporal sampling.}
Tab.~\ref{tab:app:temporal} reports the trajectory sampling settings of
pretraining. The effective sampling rate $f$ accounts for temporal
subsampling: $H$ future intervals span $H/f$ seconds and yield $H+1$
coordinates including the current anchor. We observe slower robot motion
in AgiBot and RoboMIND and therefore
subsample their 45-Hz trajectories with stride 3 (15 Hz). This extends
the physical time covered by each action chunk, bringing its spatial
motion extent closer to that of chunks from the other datasets.
VITRA uses 6 Hz for Ego4D, 8 Hz for EgoExo4D and Something-Something V2,
and 10 Hz for EPIC-KITCHENS. $H_{\mathrm{hist}}$ specifies the data history
window.

\begin{table}[!htbp]
    \centering
    \small
    \setlength{\tabcolsep}{5pt}
    \renewcommand{\arraystretch}{1.12}
    \caption{Pretraining trajectory sampling configuration. $H$ and
    $H_{\mathrm{hist}}$ count future and historical intervals; duration is
    $H/f$ in seconds. $H+1$ includes the current anchor.}
    \label{tab:app:temporal}
    \vspace{3pt}
    \begin{tabular}{@{}llccccc@{}}
        \toprule
        Dataset & Source & $f$ (Hz) & $H$ & $H_{\mathrm{hist}}$ & Duration & $H+1$ \\
        \midrule
        RoboTwin & Simulation & 15 & 20 & 20 & 1.33 & 21 \\
        LIBERO & Simulation & 20 & 20 & 20 & 1.00 & 21 \\
        VLABench & Simulation & 10 & 15 & 15 & 1.50 & 16 \\
        DROID & Real robot & 15 & 20 & 20 & 1.33 & 21 \\
        AgiBot & Real robot & 15 & 20 & 20 & 1.33 & 21 \\
        RoboMIND & Real robot & 15 & 20 & 20 & 1.33 & 21 \\
        RH20T & Real robot & 20 & 20 & 20 & 1.00 & 21 \\
        W1-data & Real robot & 15 & 20 & 20 & 1.33 & 21 \\
        VITRA & Human hand & 6/8/10 & 10 & 10 & $10/f$ & 11 \\
        \bottomrule
    \end{tabular}
\end{table}

\subsection{Physical-Time Trajectory Tokenizer}
\label{app:trajectory_tokenizer}

The encoder and decoder are defined in
Eq.~\eqref{eq:physical_time_flowchunk}.
The temporal basis takes timestamps in seconds on the shared scale
$T_{\max}=3\,\mathrm{s}$:
\begin{equation}
    [\boldsymbol\phi_{\mathrm{time}}(\tau)]_j
    =\operatorname{sgn}(\tau)\,
    \sin\!\left[\frac{\pi}{2}
    \left|\frac{\tau}{T_{\max}}\right|^{j/d_z}\right],
    \qquad j=0,\ldots,d_z-1,
\label{eq:tokenizer_time_basis}
\end{equation}
with latent dimension $d_z=256$.
Future trajectories use positive offsets; history uses negative offsets
and displacements from past positions toward the current anchor.
An input MLP produces $\mathbf h_{n,k}$ from $\Delta\mathbf x_{n,k}$, and
separate temporal MLPs produce the encoder and decoder FiLM parameters.

\paragraph{Reconstruction training.}
We minimize squared Euclidean error over valid point--time pairs,
averaging the future and historical action losses and adding available
scene reconstruction loss. The trained tokenizer is frozen, and its
latents are standardized with fixed channelwise training statistics
before world-action model training.

\paragraph{Spatial anchoring and query sets.}
The spatial anchor is separate from the displacement latent. Pointwise
Fourier features of $\mathbf x_{t,n}$ are projected to the point dynamics expert's
hidden dimension and added to the trajectory token throughout denoising.
Shared token processing and order-independent attention masks make
prediction permutation equivariant within each query type: reordering
query points and their noise tokens together reorders the outputs.
Variable query counts support sparse object-level prediction or denser
scene coverage.

\paragraph{Held-out trajectory reconstruction.}
We evaluate the frozen tokenizer on 600 held-out trajectory windows per
source and trajectory type, none used for tokenizer training. Each point
trajectory is encoded into one token and decoded at its original physical
timestamps. Tab.~\ref{tab:app:tokenizer_reconstruction} reports ADE,
the mean Euclidean error across future timesteps, and FDE, the error at
the final timestep, averaged over point trajectories. The millimeter-scale
errors indicate that the tokenizer preserves motion information across
robot action flow, human-hand action flow, and scene flow in these samples.

\begin{center}
\begin{minipage}{0.805\linewidth}
    \makeatletter\def\@captype{table}\makeatother
    \centering
    \small
    \setlength{\tabcolsep}{6pt}
    \renewcommand{\arraystretch}{1.12}
    \caption{Held-out trajectory reconstruction errors of the frozen
    \mbox{tokenizer} (mm). Dashes indicate unavailable scene-flow ground truth.}
    \label{tab:app:tokenizer_reconstruction}
    \begin{tabular*}{\linewidth}{@{\extracolsep{\fill}}lcccc@{}}
        \toprule
        \multirow{2}{*}{Data source}
        & \multicolumn{2}{c}{Action flow} & \multicolumn{2}{c}{Scene flow} \\
        \cmidrule(lr){2-3}\cmidrule(lr){4-5}
        & ADE $\downarrow$ & FDE $\downarrow$
        & ADE $\downarrow$ & FDE $\downarrow$ \\
        \midrule
        RoboTwin & 1.96 & 4.62 & 1.01 & 2.32 \\
        LIBERO & 2.54 & 6.05 & 0.78 & 1.56 \\
        VLABench & 3.38 & 6.25 & 0.94 & 1.89 \\
        DROID & 3.17 & 7.85 & -- & -- \\
        AgiBot & 2.26 & 4.62 & -- & -- \\
        RoboMIND & 3.48 & 7.60 & -- & -- \\
        RH20T & 3.16 & 4.91 & -- & -- \\
        W1 & 1.70 & 3.33 & -- & -- \\
        VITRA (human hand) & 4.68 & 8.18 & -- & -- \\
        \bottomrule
    \end{tabular*}
\end{minipage}
\end{center}

\noindent\begin{minipage}{\linewidth}
\subsection{Architecture and Optimization}
\label{app:implementation}

Tab.~\ref{tab:app:architecture} summarizes the model configuration.

\begin{center}
    \makeatletter\def\@captype{table}\makeatother
    \centering
    \small
    \setlength{\tabcolsep}{6pt}
    \renewcommand{\arraystretch}{1.12}
    \caption{Model architecture. Parameter counts include the frozen tokenizer.}
    \label{tab:app:architecture}
    \begin{tabular}{@{}llr@{}}
        \toprule
        Component & Configuration & Params. (M) \\
        \midrule
        VLM & Florence-2-large & 570.7 \\
        Point dynamics expert & 28 layers; width 1,280; 20 heads & 826.3 \\
        Action expert & 28 layers; width 640; 10 heads & 256.0 \\
        Trajectory tokenizer & Time-conditioned MLPs; latent dim. 256 & 1.9 \\
        Embeddings and projections & Spatial and modality interfaces & 39.7 \\
        \midrule
        Total & & 1,694.5 \\
        \bottomrule
    \end{tabular}
\end{center}
\end{minipage}

\paragraph{Token structure and attention.}
The mask is specified in Sec.~\ref{sec:method:architecture}.
The action expert has the same depth as the dynamics expert and a smaller
hidden dimension, with one noisy command token per future timestep.
Physical execution time $\tau_k$ enters command tokens through FiLM,
separately from the denoising time.

\paragraph{Flow matching and inference.}
The interpolation and the loss weights are given in
Sec.~\ref{sec:method:architecture}.
At inference, action flow and commands use the same Euler schedule for
10 steps. Scene denoising starts from independent noise.

\paragraph{Training configuration.}
We pretrain for 200K steps with a total batch size of 1,024 using AdamW.
The learning rates are $10^{-4}$ for the point dynamics and action experts and
$10^{-5}$ for the VLM, with weight decay $10^{-8}$.
We use constant learning rates after 1,000 steps of linear warmup.
Training uses 64 BW1000\_H DCUs (64\,GiB each) with FSDP.

\paragraph{Queries and objectives.}
Robot grippers use 20 query points sampled from URDF-defined gripper geometry
per arm; human hands use 21 MANO-derived keypoints per observed hand.
Scene points are voxelized and capped at 12,000 before random
subsampling. The mixed-dataset run retains 1\% (approximately 120 scene
queries at the cap), with 5-mm voxels for the simulation datasets.
RGB inputs are resized to $224\!\times\!224$.
The flow-matching weights are set in
Sec.~\ref{sec:method:architecture}; unavailable supervision is masked.
This run uses a frustum-routing substitution probability of 0.1. Action position coordinates are scaled by 10; rotation and gripper
coordinates retain unit scale. Inference uses 10 Euler steps for each
active expert.

\subsection{Inference Efficiency}
\label{app:inference_efficiency}

We measure inference latency on a single NVIDIA GeForce RTX 5090 D.
As shown in Tab.~\ref{tab:app:inference_efficiency}, \method{} directly
produces executable actions without generating action or scene flows
in 55.77\,ms per chunk, providing a $2.20\times$ speedup over jointly
generating actions and action flow. With only 1.7B parameters, this mode
is $2.81\times$ faster than Fast-WAM without video generation and
substantially faster than the joint video--action baselines
Fast-WAM~\citep{yuan2026fastwam} and Motus~\citep{bi2025motus},
supporting low-latency robot control. In world-model mode, the number
of scene queries can be adjusted at inference to match the desired
spatial sampling density and compute budget. Reducing scene queries
from 3,000 to 1,024 lowers joint prediction latency from 561.60\,ms
to 305.49\,ms, allowing a flexible trade-off between scene coverage
and inference cost.

\begin{table}[!htbp]
    \centering
    \small
    \setlength{\tabcolsep}{6pt}
    \renewcommand{\arraystretch}{1.08}
    \caption{\textbf{Inference efficiency on an RTX 5090 D.}
    Latency is measured per prediction chunk; $N_s$ denotes the number
    of scene queries.}
    \label{tab:app:inference_efficiency}
    \begin{tabular}{@{}lrrr@{}}
        \toprule
        Method / output mode & Params (B) & $N_s$ & Latency (ms) \\
        \midrule
        Fast-WAM (w/o video) & 6.0 & -- & 156.81 \\
        Fast-WAM (joint) & 6.0 & -- & 579.79 \\
        Motus & 8.0 & -- & 2,414.90 \\
        \midrule
        \method{}: actions only & 1.7 & -- & 55.77 \\
        \method{}: actions + action flow & 1.7 & -- & 122.51 \\
        \method{}: actions + both flows & 1.7 & 1,024 & 305.49 \\
        \method{}: actions + both flows & 1.7 & 3,000 & 561.60 \\
        \bottomrule
    \end{tabular}
\end{table}

\section{Spatial Grounding without Depth Input}
\label{app:missing_depth}

Candidate depths, the router, and $\mathcal L_{\mathrm{PE}}$ are defined
in Sec.~\ref{sec:method:observation},
Eqs.~\eqref{eq:frustum_routing} and~\eqref{eq:position_distillation}.
During training, routed features occasionally replace depth-derived
features on valid regions, with the substitution rate reported in
Appendix~\ref{app:implementation}.
This section evaluates the router when depth is removed at test time.

\subsection{Evaluation without Depth Input}
\label{app:missing_depth:evaluation}

We evaluate policy execution on LIBERO-Plus at 100K training steps,
without mixed-dataset pretraining or scene-flow supervision.
Tab.~\ref{tab:app:missing_depth} compares routed PE without depth input
against ground-truth depth PE using the same checkpoint, and includes
the no-3D-PE baseline from Tab.~\ref{tab:experiments:libero_plus_ablation}.
Camera calibration and robot-state inputs remain available in all settings.

\begin{table}[!htbp]
    \centering
    \footnotesize
    \setlength{\tabcolsep}{3pt}
    \renewcommand{\arraystretch}{1.15}
    \caption{\textbf{Spatial grounding without depth input on LIBERO-Plus.}
    Success rates (\%, $\uparrow$) over 10,030 episodes. Routed PE and
    ground-truth depth PE use the same checkpoint.}
    \label{tab:app:missing_depth}
    \begin{adjustbox}{max width=\linewidth}
    \begin{tabular}{@{}lrrrrrrrr@{}}
        \toprule
        Visual spatial encoding
        & Camera & Robot & Language & Light & Background & Noise & Layout & Overall \\
        \midrule
        No visual 3D PE
        & 66.73 & 79.29 & 75.21 & 94.57 & 92.19 & 84.13 & 84.46 & 81.35 \\
        Routed PE (no depth)
        & 83.74 & 79.61 & 74.30 & 91.68 & 84.20 & 93.38 & 85.57 & 84.43 \\
        \midrule
        Ground-truth depth PE
        & 84.37 & 79.68 & 74.24 & 91.94 & 85.13 & 93.88 & 85.18 & 84.68 \\
        \bottomrule
    \end{tabular}
    \end{adjustbox}
\end{table}

\paragraph{Results.}
Without depth input, routed PE achieves 84.43\% overall success,
closely matching the 84.68\% obtained with ground-truth depth PE.
The gap of only 0.25 percentage points supports the effectiveness of
distilling depth-derived spatial information into RGB-conditioned
positional features, preserving policy performance when depth is
unavailable at inference. Compared with removing visual 3D PE, routed
PE improves overall success by 3.08 points, with the largest gains
under camera-viewpoint perturbations and sensor noise.

\section{Detailed LIBERO-Plus Results}
\label{app:libero_plus}

For both training settings, we train \method{} on standard LIBERO demonstrations
and use the same checkpoints as in the LIBERO evaluation to evaluate directly
on LIBERO-Plus. No additional training or adaptation is performed on
LIBERO-Plus; this is an out-of-distribution evaluation.

Tab.~\ref{tab:experiments:libero_plus} breaks down success rates across
all seven perturbation types in LIBERO-Plus~\citep{fei2025liberoplus};
Tab.~\ref{tab:experiments:simulation} reports overall success.
The comparison additionally includes OpenVLA~\citep{kim2024openvla}
and ABot-M0.5~\citep{chen2026abotm05}.

%

\begin{table}[!htbp]
    \centering
    \small
    \setlength{\tabcolsep}{3pt}
    \renewcommand{\arraystretch}{1.10}
    \caption{LIBERO-Plus success rates (\%, $\uparrow$) across seven
    perturbation types. Robot denotes initial-state perturbations.}
    \label{tab:experiments:libero_plus}
    \vspace{3pt}
    \begin{adjustbox}{max width=\linewidth}
    \begin{tabular}{@{}lccccccccc@{}}
        \toprule
        Method & Params (B) & Camera & Robot & Language & Light
        & Background & Noise & Layout & Overall \\
        \midrule
        X-VLA & 1 & 23.4 & \underline{89.7} & 75.7 & 88.2 & 96.0 & 62.7 & 71.8 & 71.4 \\
        $\pi_{0}$ & 3 & 13.8 & 6.0 & 58.8 & 85.0 & 81.4 & 79.0 & 68.9 & 53.6 \\
        $\pi_{0.5}$ & 3 & 75.4 & 77.5 & 85.6 & \underline{96.9} & 94.6 & 89.7 & \underline{85.7} & 85.7 \\
        OpenVLA & 7 & 0.8 & 3.5 & 23.0 & 8.1 & 34.8 & 15.2 & 28.5 & 15.6 \\
        Cosmos Policy & 2 & 75.8 & 63.3 & 81.7 & 96.5 & 88.9 & \textbf{92.7} & 82.2 & 82.2 \\
        LingBot-VA & 5.3 & 40.9 & 83.0 & 86.4 & 82.3 & 53.1 & 64.4 & 76.2 & 69.5 \\
        Fast-WAM & 6 & 16.4 & 44.5 & 68.9 & 78.2 & 53.7 & 37.7 & 60.7 & 51.5 \\
        ABot-M0.5 & 6 & 70.5 & 87.4 & \textbf{88.6} & 94.0 & 89.7 & 75.5 & 85.2 & 83.4 \\
        LaMP & 5 & 64.5 & 69.6 & \underline{88.2} & 95.3 & \textbf{97.4} & 76.9 & 73.8 & 79.3 \\
        \midrule
        \method{} (w/o Pretrain) & 1.7 & \underline{81.9} & 84.8 & 77.6 & 96.7 & 94.4 & \underline{89.8} & 85.1 & \underline{86.5} \\
        \method{} & 1.7 & \textbf{82.1} & \textbf{94.2} & 85.0 & \textbf{98.1} & \underline{96.3} & 84.5 & \textbf{87.7} & \textbf{89.1} \\
        \bottomrule
    \end{tabular}
    \end{adjustbox}
\end{table}

\section{Detailed VLABench Results}
\label{app:vlabench}

Tab.~\ref{tab:app:vlabench} reports five evaluation tracks:
in-distribution, cross-category, commonsense, semantic instruction, and
unseen texture (Tracks~1, 2, 3, 4, and 6). The comparison additionally
includes $\pi_0$-FAST~\citep{pertsch2025fast}.

\begin{center}
\begin{minipage}{\linewidth}
\makeatletter\def\@captype{table}\makeatother
    \centering
    \small
    \setlength{\tabcolsep}{4pt}
    \renewcommand{\arraystretch}{1.15}
    \caption{Detailed VLABench results (\%). SR, PS, and IS denote
    success rate, progress score, and intention score, respectively
    (all $\uparrow$).}
    \label{tab:app:vlabench}
    \vspace{3pt}
    \begin{adjustbox}{max width=\linewidth}
    \begin{tabular*}{0.94\linewidth}{@{\extracolsep{\fill}}l*{9}{r}@{}}
        \toprule
        Method & \multicolumn{3}{c}{In-distribution}
        & \multicolumn{3}{c}{Cross-category}
        & \multicolumn{3}{c}{Commonsense} \\
        \cmidrule(lr){2-4}\cmidrule(lr){5-7}\cmidrule(lr){8-10}
        & SR & PS & IS & SR & PS & IS & SR & PS & IS \\
        \midrule
        $\pi_0$ & 47.0 & 62.7 & 67.8 & 21.2 & 33.6 & 44.0 & 29.1 & 43.0 & 54.9 \\
        $\pi_0$-FAST & 56.2 & 66.8 & 72.4 & \underline{31.0} & 38.2 & \underline{47.8} & 38.0 & 48.6 & 56.8 \\
        X-VLA & 40.0 & 66.5 & \textbf{88.0} & 11.0 & 24.0 & 42.0 & 33.0 & 51.7 & 66.0 \\
        $\pi_{0.5}$ & \underline{65.4} & \underline{77.8} & 80.4 & \textbf{38.2} & \textbf{49.7} & \textbf{52.0} & 43.9 & 57.3 & 60.0 \\
        \midrule
        \method{} (w/o Pretrain) & 62.0 & 76.5 & 80.2 & 25.4 & 37.2 & 45.7 & \underline{49.0} & \underline{63.4} & \textbf{72.8} \\
        \method{} & \textbf{75.1} & \textbf{85.6} & \underline{82.3} & 28.9 & \underline{39.1} & 44.4 & \textbf{53.8} & \textbf{65.9} & \underline{68.5} \\
        \midrule
        Method & \multicolumn{3}{c}{Semantic instruction}
        & \multicolumn{3}{c}{Unseen texture}
        & \multicolumn{3}{c}{Avg.} \\
        \cmidrule(lr){2-4}\cmidrule(lr){5-7}\cmidrule(lr){8-10}
        & SR & PS & IS & SR & PS & IS & SR & PS & IS \\
        \midrule
        $\pi_0$ & 17.3 & 38.7 & 58.0 & 32.2 & 42.5 & 50.6 & 29.4 & 44.1 & 55.0 \\
        $\pi_0$-FAST & 35.0 & 45.0 & 59.4 & 39.0 & 49.0 & 56.8 & 39.8 & 49.5 & 58.6 \\
        X-VLA & 39.0 & 62.5 & \textbf{82.0} & 24.0 & 51.2 & \underline{73.0} & 29.4 & 51.2 & \underline{70.2} \\
        $\pi_{0.5}$ & 48.2 & 64.2 & 67.0 & 44.9 & 62.3 & 65.0 & 48.1 & 62.3 & 64.9 \\
        \midrule
        \method{} (w/o Pretrain) & \underline{62.4} & \underline{75.8} & \underline{79.8} & \underline{52.6} & \textbf{68.5} & \textbf{77.6} & \underline{50.3} & \underline{64.3} & \textbf{71.2} \\
        \method{} & \textbf{67.7} & \textbf{79.9} & 78.7 & \textbf{55.4} & \underline{67.4} & 72.7 & \textbf{56.2} & \textbf{67.6} & 69.3 \\
        \bottomrule
    \end{tabular*}
    \end{adjustbox}
\end{minipage}
\end{center}

\section{Detailed RoboTwin Results}
\label{app:robotwin}

Tab.~\ref{tab:app:robotwin} gives per-task results for both training settings
in Tab.~\ref{tab:experiments:simulation}. Each setting contains 50 tasks
with 100 evaluation episodes per task; averages weight all tasks equally.

\begin{center}
\begin{minipage}{\linewidth}
\makeatletter\def\@captype{table}\makeatother
    \centering
    \fontsize{8}{9}\selectfont
    \setlength{\tabcolsep}{3pt}
    \renewcommand{\arraystretch}{0.98}
    \caption{Detailed RoboTwin 2.0 success rates (\%, $\uparrow$).
    Clean and Rand.\ denote \texttt{demo\_clean} and \texttt{demo\_randomized}.
    All methods are evaluated over 100 episodes per task in each setting.}
    \label{tab:app:robotwin}
    \vspace{3pt}
    \begin{tabular*}{\linewidth}{@{\extracolsep{\fill}}l*{10}{c}@{}}
        \toprule
        Task & \multicolumn{2}{c}{$\pi_{0.5}$}
        & \multicolumn{2}{c}{Motus} & \multicolumn{2}{c}{Fast-WAM}
        & \multicolumn{2}{c}{\shortstack{\method{}\\(w/o Pretrain)}}
        & \multicolumn{2}{c}{\method{}} \\
        \cmidrule(lr){2-3}\cmidrule(lr){4-5}\cmidrule(lr){6-7}
        \cmidrule(lr){8-9}\cmidrule(lr){10-11}
        & Clean & Rand. & Clean & Rand. & Clean & Rand.
        & Clean & Rand. & Clean & Rand. \\
        \midrule
        Adjust Bottle & \textbf{100} & \underline{99} & \underline{89} & 93 & \textbf{100} & \textbf{100} & \textbf{100} & 98 & \textbf{100} & \textbf{100} \\
        Beat Block Hammer & 96 & 93 & 95 & 88 & \underline{99} & \underline{97} & 97 & 92 & \textbf{100} & \textbf{98} \\
        Blocks Ranking RGB & 92 & 85 & \underline{99} & 97 & \textbf{100} & \textbf{100} & \textbf{100} & \underline{99} & 97 & 94 \\
        Blocks Ranking Size & 49 & 26 & 75 & 63 & \textbf{94} & \textbf{98} & \underline{91} & \underline{83} & 78 & 82 \\
        Click Alarmclock & \underline{98} & \underline{89} & \textbf{100} & \textbf{100} & \textbf{100} & \textbf{100} & \textbf{100} & \textbf{100} & \textbf{100} & \textbf{100} \\
        Click Bell & \underline{99} & \underline{66} & \textbf{100} & \textbf{100} & \textbf{100} & \textbf{100} & \textbf{100} & \textbf{100} & \textbf{100} & \textbf{100} \\
        Dump Bin Bigbin & 92 & \underline{97} & 95 & 91 & \underline{97} & 96 & \underline{97} & \textbf{98} & \textbf{98} & \underline{97} \\
        Grab Roller & \textbf{100} & \textbf{100} & \textbf{100} & \textbf{100} & \textbf{100} & \textbf{100} & \textbf{100} & \textbf{100} & \textbf{100} & \textbf{100} \\
        Handover Block & 66 & 57 & 86 & 73 & \underline{95} & \underline{81} & 88 & 74 & \textbf{96} & \textbf{84} \\
        Handover Mic & 98 & 97 & 78 & 63 & \underline{99} & \textbf{100} & \textbf{100} & \underline{99} & \textbf{100} & 97 \\
        Hanging Mug & 18 & 17 & 38 & 38 & \underline{58} & \textbf{62} & 48 & \underline{57} & \textbf{61} & 41 \\
        Lift Pot & 96 & 85 & 96 & \underline{99} & \textbf{100} & \textbf{100} & 98 & \underline{99} & \underline{99} & \underline{99} \\
        Move Can Pot & 51 & 55 & 34 & 74 & 90 & 88 & \textbf{97} & \textbf{100} & \underline{94} & \underline{96} \\
        Move Pillbottle Pad & 84 & 61 & 93 & 96 & \textbf{100} & \textbf{99} & \underline{99} & \textbf{99} & 93 & \underline{98} \\
        Move Playingcard Away & 96 & 84 & \textbf{100} & \underline{96} & \textbf{100} & \textbf{100} & \underline{98} & \textbf{100} & \textbf{100} & \textbf{100} \\
        Move Stapler Pad & 56 & 42 & \textbf{83} & \textbf{85} & 77 & 64 & 54 & 50 & \underline{79} & \underline{78} \\
        Open Laptop & 90 & 96 & 95 & 91 & \underline{98} & \textbf{100} & \underline{98} & \underline{99} & \textbf{100} & \textbf{100} \\
        Open Microwave & 34 & 77 & \underline{95} & \textbf{91} & 62 & 45 & \textbf{96} & \underline{86} & 74 & 80 \\
        Pick Diverse Bottles & 81 & 71 & \textbf{90} & \textbf{91} & 80 & 85 & 84 & 81 & \underline{86} & \underline{88} \\
        Pick Dual Bottles & 93 & 63 & \underline{96} & 90 & \textbf{100} & \underline{96} & \underline{96} & \textbf{98} & \textbf{100} & \textbf{98} \\
        Place A2B Left & 87 & 82 & 88 & 79 & \underline{95} & 93 & 90 & \underline{95} & \textbf{100} & \textbf{98} \\
        Place A2B Right & 87 & 84 & 91 & 87 & \textbf{93} & \textbf{99} & \underline{92} & 95 & \textbf{93} & \underline{98} \\
        Place Bread Basket & 77 & 64 & \underline{91} & \underline{94} & \underline{91} & 93 & \textbf{95} & \textbf{98} & \textbf{95} & 91 \\
        Place Bread Skillet & 85 & 66 & 86 & 83 & \textbf{90} & \underline{93} & 87 & \textbf{95} & \underline{89} & 86 \\
        Place Burger Fries & 94 & 87 & \underline{98} & \underline{98} & 96 & \textbf{99} & \textbf{100} & \textbf{99} & 96 & \underline{98} \\
        Place Can Basket & 62 & 62 & \textbf{81} & \underline{76} & 71 & 69 & 72 & 74 & \underline{77} & \textbf{92} \\
        Place Cans Plasticbox & 94 & 84 & \underline{98} & 94 & \textbf{99} & 96 & \textbf{99} & \underline{97} & \textbf{99} & \textbf{99} \\
        Place Container Plate & \underline{99} & 95 & 98 & \underline{99} & 96 & \textbf{100} & \underline{99} & 95 & \textbf{100} & \underline{99} \\
        Place Dual Shoes & 75 & 75 & 93 & 87 & \underline{94} & 88 & 92 & \underline{91} & \textbf{99} & \textbf{96} \\
        Place Empty Cup & \textbf{100} & \underline{99} & \underline{99} & 98 & \textbf{100} & \textbf{100} & \textbf{100} & \textbf{100} & \textbf{100} & \textbf{100} \\
        Place Fan & 87 & 85 & \underline{91} & 87 & \textbf{96} & \textbf{96} & 89 & 92 & 90 & \underline{94} \\
        Place Mouse Pad & 60 & 39 & 66 & 68 & 83 & \underline{89} & \underline{89} & \textbf{92} & \textbf{91} & 87 \\
        Place Object Basket & 80 & 76 & 81 & 87 & \underline{89} & 88 & 79 & \textbf{94} & \textbf{94} & \underline{91} \\
        Place Object Scale & 86 & 80 & 88 & 85 & \underline{90} & \textbf{97} & \textbf{91} & \underline{94} & \textbf{91} & 91 \\
        Place Object Stand & 91 & 85 & \textbf{98} & \textbf{97} & 90 & \underline{94} & 92 & 88 & \underline{97} & \textbf{97} \\
        Place Phone Stand & 81 & 81 & 87 & 86 & \textbf{97} & \textbf{99} & 85 & 89 & \underline{89} & \underline{95} \\
        Place Shoe & 92 & 93 & \underline{99} & 97 & 96 & \textbf{99} & 97 & 97 & \textbf{100} & \underline{98} \\
        Press Stapler & 87 & 83 & \underline{93} & \textbf{98} & 90 & \underline{97} & \textbf{98} & \textbf{98} & \textbf{98} & 96 \\
        Put Bottles Dustbin & 84 & 79 & 81 & 79 & \textbf{95} & \underline{90} & \underline{87} & \underline{90} & 85 & \textbf{98} \\
        Put Object Cabinet & 80 & 79 & \underline{88} & 71 & \textbf{94} & \underline{89} & \underline{88} & 84 & 77 & \textbf{92} \\
        Rotate QR Code & 89 & 87 & 89 & 73 & \textbf{93} & 89 & 91 & \underline{91} & \underline{92} & \textbf{92} \\
        Scan Object & 72 & 65 & 67 & 66 & \underline{89} & \underline{92} & 87 & \underline{92} & \textbf{97} & \textbf{96} \\
        Shake Bottle & \underline{99} & 97 & \textbf{100} & 97 & \textbf{100} & \textbf{100} & \textbf{100} & 97 & \textbf{100} & \underline{99} \\
        Shake Bottle Horizontally & \underline{99} & \underline{99} & \textbf{100} & 98 & \textbf{100} & \textbf{100} & \textbf{100} & 94 & \textbf{100} & \textbf{100} \\
        Stack Blocks Three & 91 & 76 & 91 & \underline{95} & \underline{95} & \textbf{97} & 93 & 84 & \textbf{98} & 94 \\
        Stack Blocks Two & 97 & \textbf{100} & \textbf{100} & \underline{98} & \textbf{100} & \textbf{100} & \underline{99} & \underline{98} & \textbf{100} & 96 \\
        Stack Bowls Three & 77 & 71 & 79 & \textbf{87} & \underline{80} & 81 & \textbf{84} & \textbf{87} & \textbf{84} & \underline{85} \\
        Stack Bowls Two & \underline{95} & 96 & \textbf{98} & 98 & 92 & 98 & 91 & \underline{99} & \underline{95} & \textbf{100} \\
        Stamp Seal & 79 & 55 & \textbf{93} & \underline{92} & \underline{90} & \textbf{94} & 79 & 63 & 83 & \underline{92} \\
        Turn Switch & 62 & 54 & \textbf{84} & \textbf{78} & 61 & 59 & 68 & 63 & \underline{83} & \underline{72} \\
        \midrule
        Average & 82.74 & 76.76 & 88.66 & 87.02 & \underline{91.88} & \underline{91.78} & 91.08 & 90.74 & \textbf{92.94} & \textbf{93.04} \\
        \bottomrule
    \end{tabular*}
\end{minipage}
\end{center}

\section{World Model Evaluation Details}
\label{app:world_model_protocol}

\subsection{Action-Conditioned Scene Prediction}
\label{app:matched_action}

\paragraph{Evaluation protocol.}
Tab.~\ref{tab:experiments:world_model} evaluates 550 test clips from
50 RoboTwin tasks, with one Clean and ten Randomized episodes per task.
Each episode contributes a fixed window near its midpoint, containing
the current frame and 20 future recorded frames. Initial scene points
are sampled with a 1.5\,cm voxel grid.
All methods share initial queries, point identities, ground-truth
trajectories, and robot-motion conditions.
For video methods, queries are projected into the initial image using camera calibration,
accounting for resizing and cropping; only in-frame projections are
tracked by SpatialTrackerV2. Ctrl-World uses Depth Anything 3 depth,
whereas X-WAM uses its native depth predictions. Depth scale for
Ctrl-World and inverse-depth affine calibration for X-WAM are estimated
from initial observed depth and fixed throughout the video. Tracked
trajectories are transformed into world coordinates, and their predicted
displacements are anchored at the shared initial queries. No future
ground-truth depth or trajectories are used for alignment.

\paragraph{Scoring.}
In our evaluation, we label a point as moving when its maximum
ground-truth displacement from its initial position over the 20 future
frames exceeds 1\,cm, and as static otherwise. These labels are fixed
per clip and shared across methods. We score future frames $4,8,12,16,20$: ADE averages Euclidean
errors over valid point--frame pairs in each group, Moving FDE measures
moving-point error at frame 20, and Moving Acc@1cm is the fraction of
valid moving-point--frame pairs with prediction error at most 1\,cm.
Metrics are first computed per clip, averaged within each task--condition
group, and then averaged equally across groups. Clips without valid
moving points are excluded only from the corresponding moving metrics.

\subsection{Closed-Loop Action--Scene Consistency}
\label{app:future_action_alignment}

We evaluate three RoboTwin tasks, Adjust Bottle, Beat Block Hammer, and
Blocks Ranking RGB, under Clean and Randomized conditions, yielding six
settings. Each method is evaluated over 10 trials per setting, for 60
trials in total. We compare with X-WAM~\citep{guo2026xwam},
Fast-WAM~\citep{yuan2026fastwam}, LingBot-VA~\citep{li2026lingbotva},
and Motus~\citep{bi2025motus}. Each generated action chunk is executed
in the simulator, and the resulting observation conditions the next
prediction. Models retain their native prediction horizons and are
scored at completed action boundaries until termination. Fast-WAM
executes all 32 actions rather than replanning after 24.

For \method{}, the point dynamics expert generates an action flow, while the action
expert separately generates an executable command chunk from the same
observation and instruction. The dynamics expert then predicts scene-flow
trajectories conditioned on its generated action flow. We compare these
predictions with the ground-truth scene trajectories resulting from
executing the action expert's command chunk in the simulator.

\paragraph{Trajectory extraction and scoring.}
Ground-truth query points follow their simulator objects or links.
SpatialTrackerV2 lifts video predictions using X-WAM's native depth or
Depth Anything 3 estimates for RGB-only models, with initial-frame depth
calibration and query anchoring. We apply the 1\,cm displacement threshold
to label moving points within each scored chunk. Metrics are averaged
within rounds, then within episodes, across trials within each setting,
and finally equally across the six settings. Rounds without valid moving points are excluded only from
moving-point metrics; all episodes are retained regardless of task outcome.

\begin{center}
\begin{minipage}{\linewidth}
\makeatletter\def\@captype{table}\makeatother
    \centering
    \small
    \setlength{\tabcolsep}{4pt}
    \renewcommand{\arraystretch}{1.10}
    \caption{\textbf{Closed-loop action--scene consistency on RoboTwin.}
    Predictions are compared with the scene evolution produced by executing
    the same model's generated actions. Errors are in cm; Moving Acc@1cm
    is in \%.}
    \label{tab:app:joint_consistency}
    \begin{adjustbox}{max width=\linewidth}
    \begin{tabular}{@{}lcccc@{}}
        \toprule
        Method & Moving ADE $\downarrow$ & Static ADE $\downarrow$
        & Moving FDE $\downarrow$ & Moving Acc@1cm $\uparrow$ \\
        \midrule
        X-WAM + STv2 & \underline{2.579} & \underline{0.166} & \underline{4.056} & 35.55 \\
        Fast-WAM + STv2 & 3.554 & 0.586 & 5.133 & 30.37 \\
        LingBot-VA + STv2 & 2.603 & 0.810 & 4.716 & 33.64 \\
        Motus + STv2 & 3.216 & 0.318 & 6.778 & \underline{35.66} \\
        \method{} & \textbf{1.211} & \textbf{0.054} & \textbf{2.020} & \textbf{64.26} \\
        \bottomrule
    \end{tabular}
    \end{adjustbox}
\end{minipage}
\end{center}

These scores measure action--scene consistency along each model's own
rollout and include video-to-3D reconstruction error for video-based methods.
For \method{}, agreement between predicted and observed scene changes also
provides indirect evidence that its generated action flow and executable
commands are consistent in their scene effects.

\section{Additional Action-Conditioned Scene Predictions}
\label{app:scene_flow_visualizations}

\subsection{Scene Prediction in Simulation}
\label{app:sim_scene_flow}

Figs.~\ref{fig:app:scene_flow_1}--\ref{fig:app:scene_flow_4} extend
Fig.~\ref{fig:experiments:scene_flow_qualitative} with 19 training clips,
showing RGB, initial points, predicted flows, and ground-truth flows.
Purple denotes conditioning gripper motion. These examples illustrate
learned geometry rather than held-out accuracy. For denser visualization,
we relax part-boundary and depth-jump filters while retaining valid-depth
and non-robot checks. Every displayed scene point is directly predicted;
initial RGB colors are retained and unobserved surfaces are not synthesized.

\begin{center}
\begin{minipage}{\linewidth}
\makeatletter\def\@captype{figure}\makeatother
    \centering
    \includegraphics[width=\linewidth]{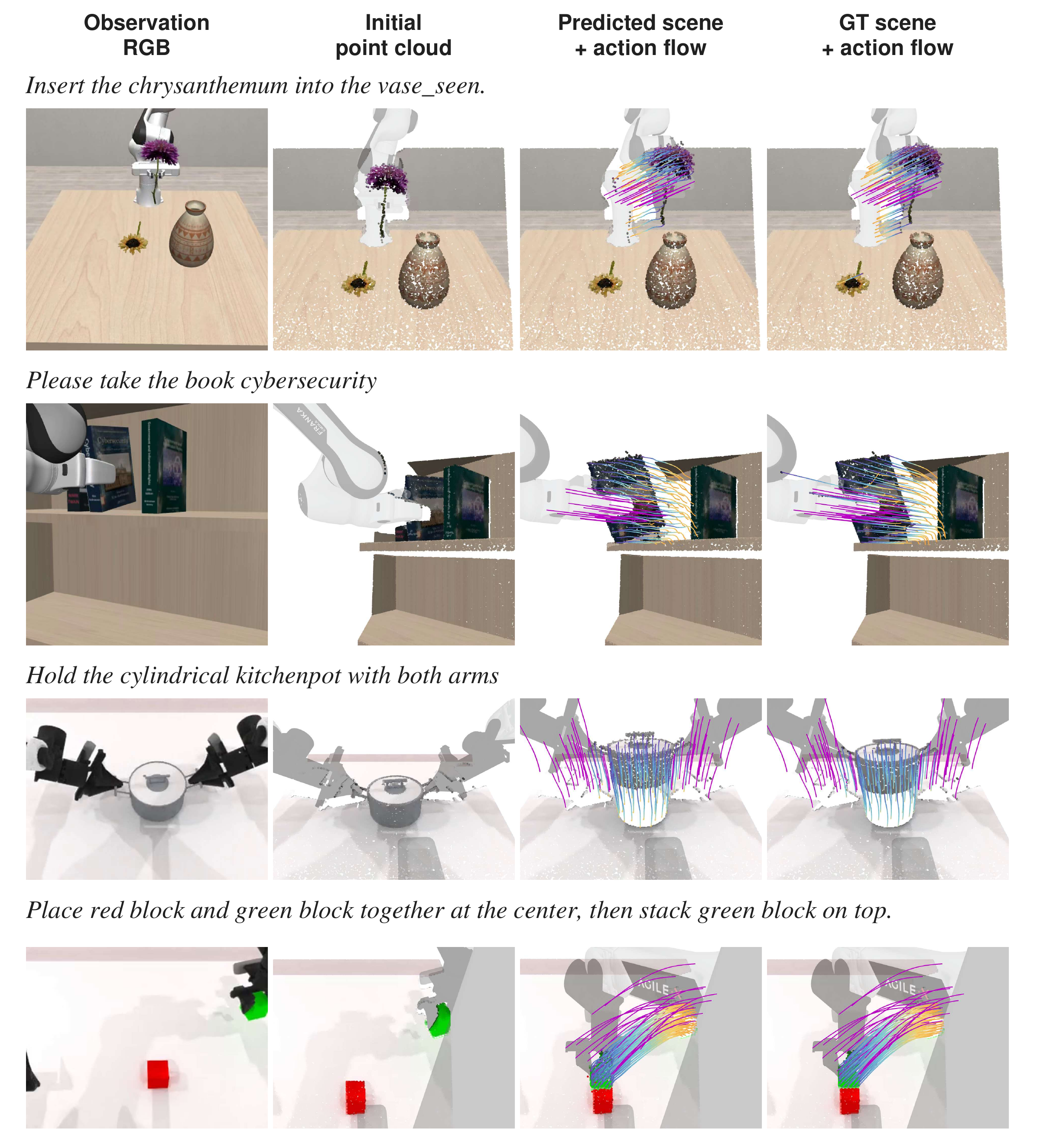}
    \caption{Additional scene-flow visualizations (1/4): insertion,
    extraction, bimanual lifting, and block stacking.}
    \label{fig:app:scene_flow_1}
\end{minipage}
\end{center}

\begin{center}
\begin{minipage}{\linewidth}
\makeatletter\def\@captype{figure}\makeatother
    \centering
    \includegraphics[width=\linewidth]{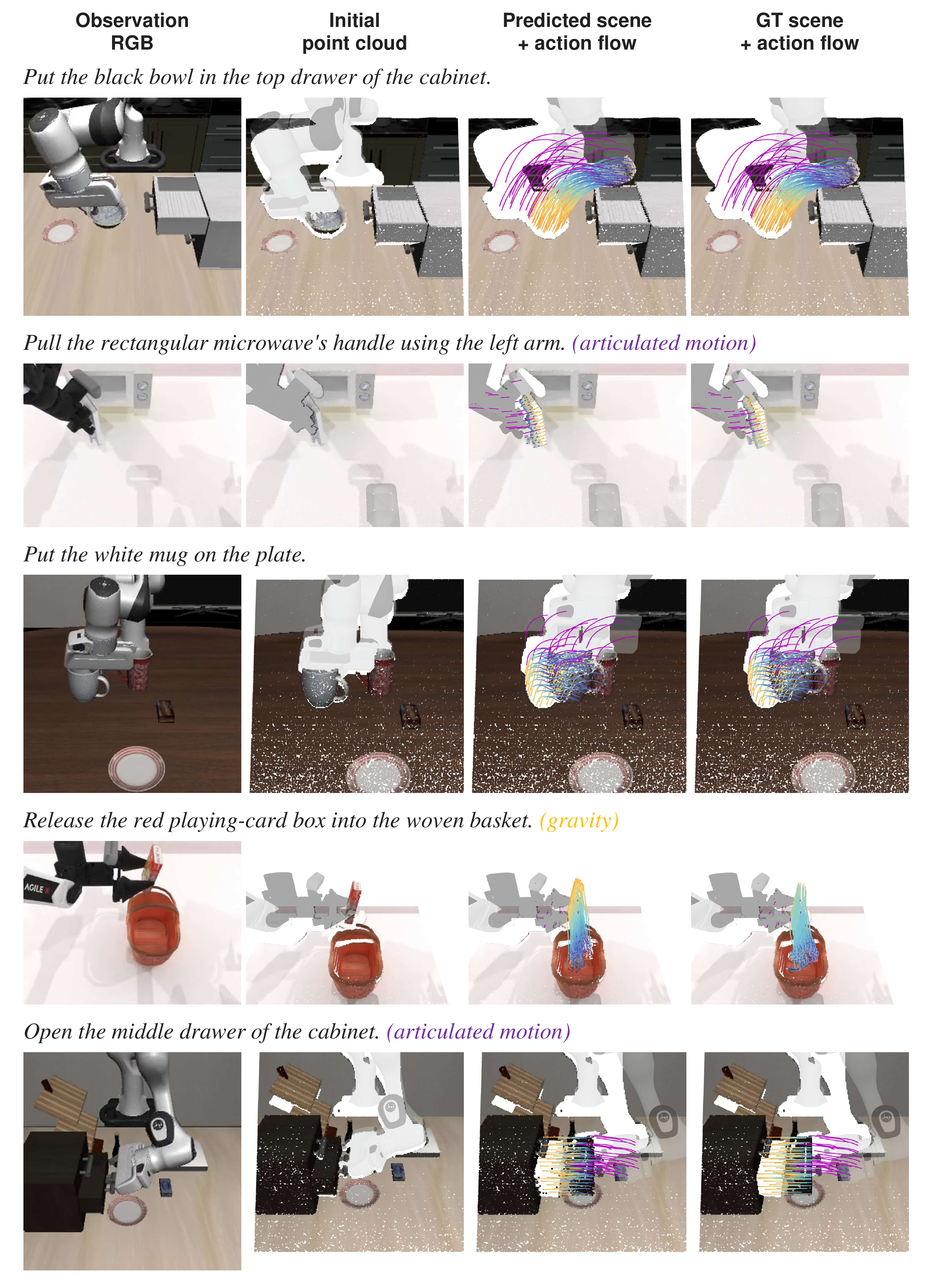}
    \caption{Additional scene-flow visualizations (2/4): bowl and mug placement,
    articulated motion, and gravity-driven motion after object release.}
    \label{fig:app:scene_flow_2}
\end{minipage}
\end{center}

\begin{center}
\begin{minipage}{\linewidth}
\makeatletter\def\@captype{figure}\makeatother
    \centering
    \includegraphics[width=\linewidth]{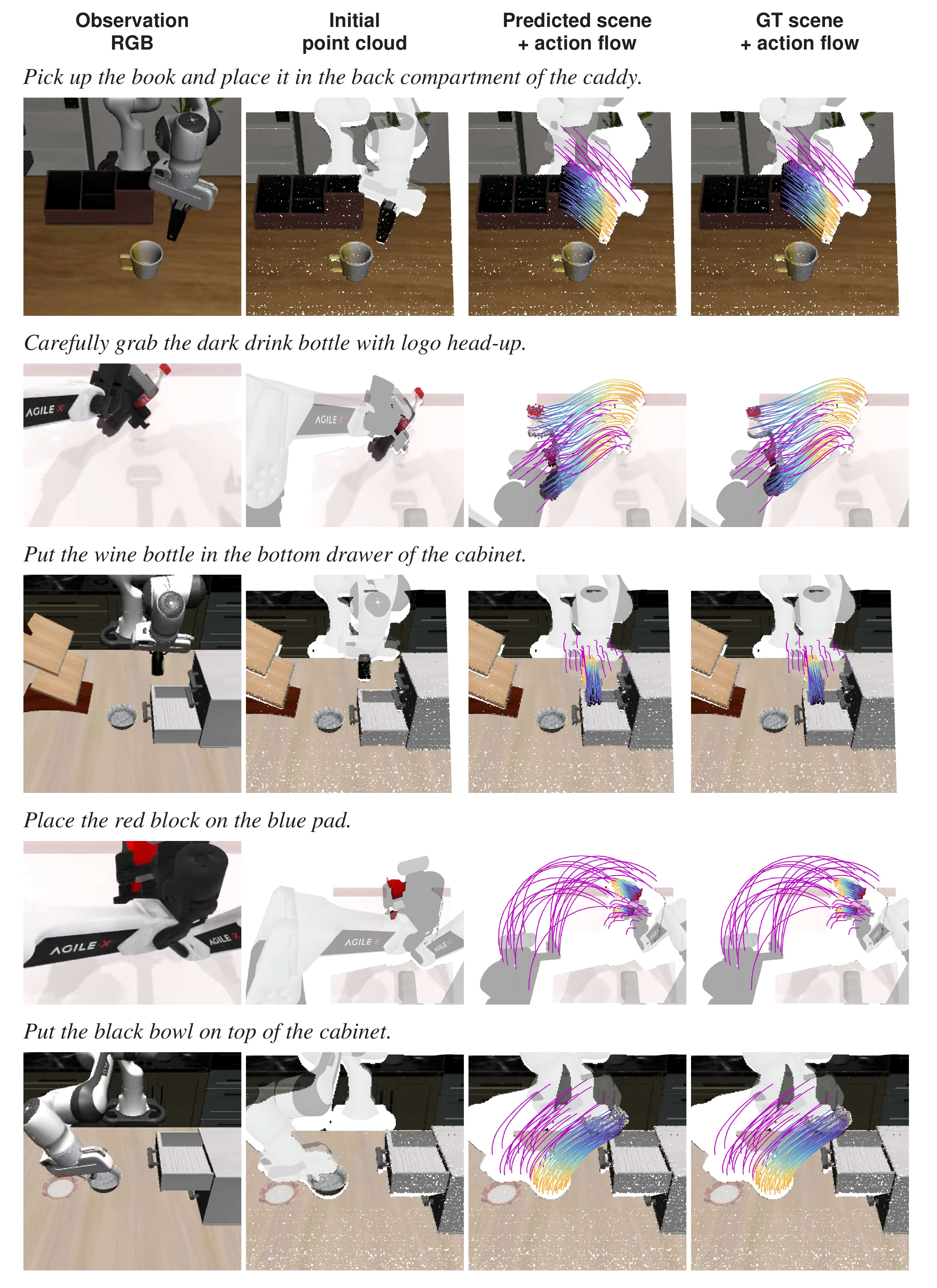}
    \caption{Additional scene-flow visualizations (3/4): book, bottle, block,
    and bowl manipulation across LIBERO and RoboTwin.}
    \label{fig:app:scene_flow_3}
\end{minipage}
\end{center}

\begin{center}
\begin{minipage}{\linewidth}
\makeatletter\def\@captype{figure}\makeatother
    \centering
    \includegraphics[width=\linewidth]{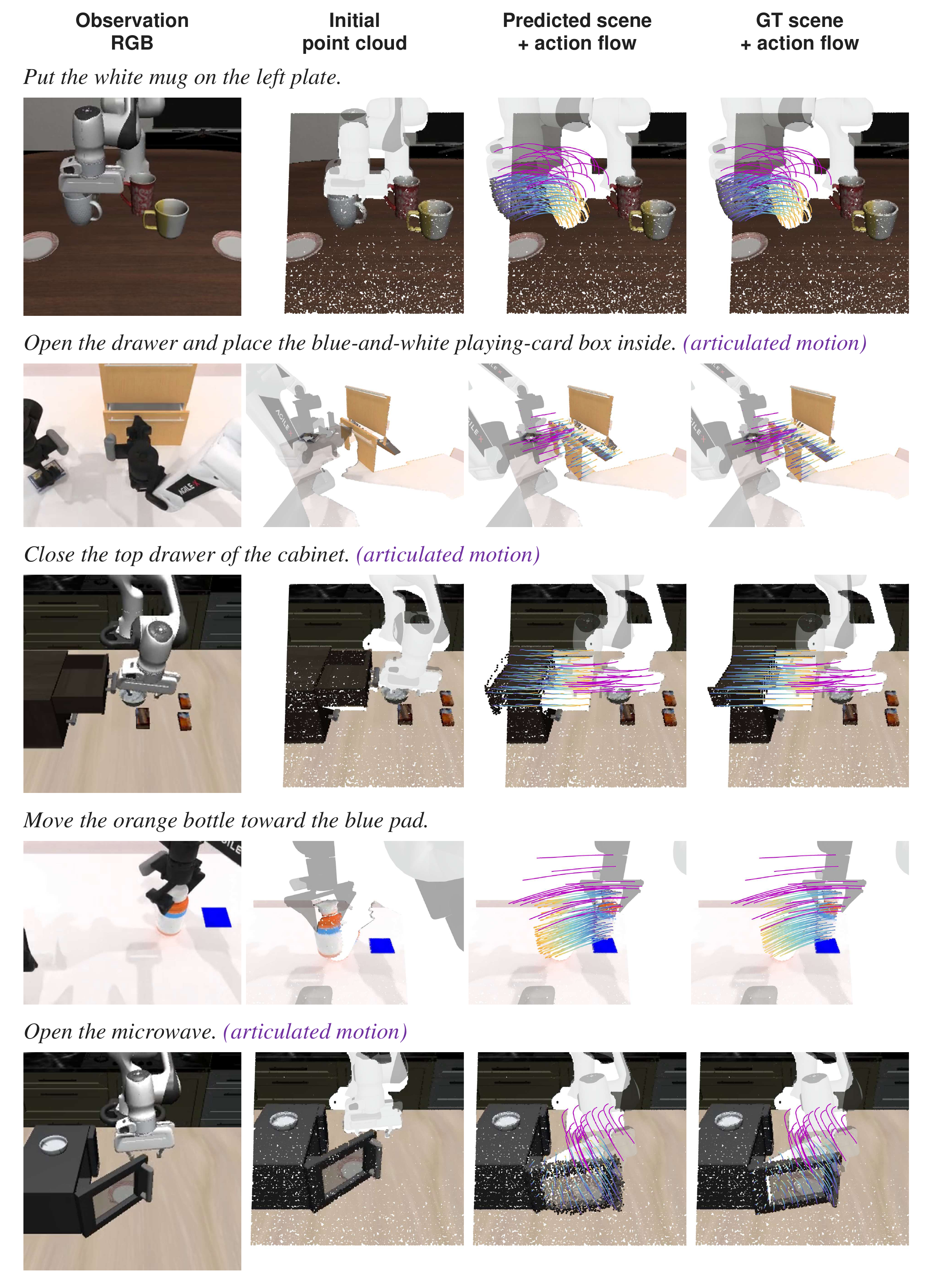}
    \caption{Additional scene-flow visualizations (4/4): object placement,
    drawer opening and closing, and microwave opening.}
    \label{fig:app:scene_flow_4}
\end{minipage}
\end{center}

\clearpage
\subsection{Transfer to Real-World Observations}
\label{app:real_scene_flow}

Fig.~\ref{fig:app:real_scene_flow} shows predictions on DROID and RH20T
conditioned on ground-truth gripper trajectories. Despite receiving
scene-flow supervision only from simulation, \method{} captures
action-conditioned scene changes in real environments outside this
supervision distribution. These qualitative results suggest that the
learned interaction dynamics can transfer from simulation to real scenes.

\begin{center}
\begin{minipage}{\linewidth}
\makeatletter\def\@captype{figure}\makeatother
    \centering
    \includegraphics[width=\linewidth]{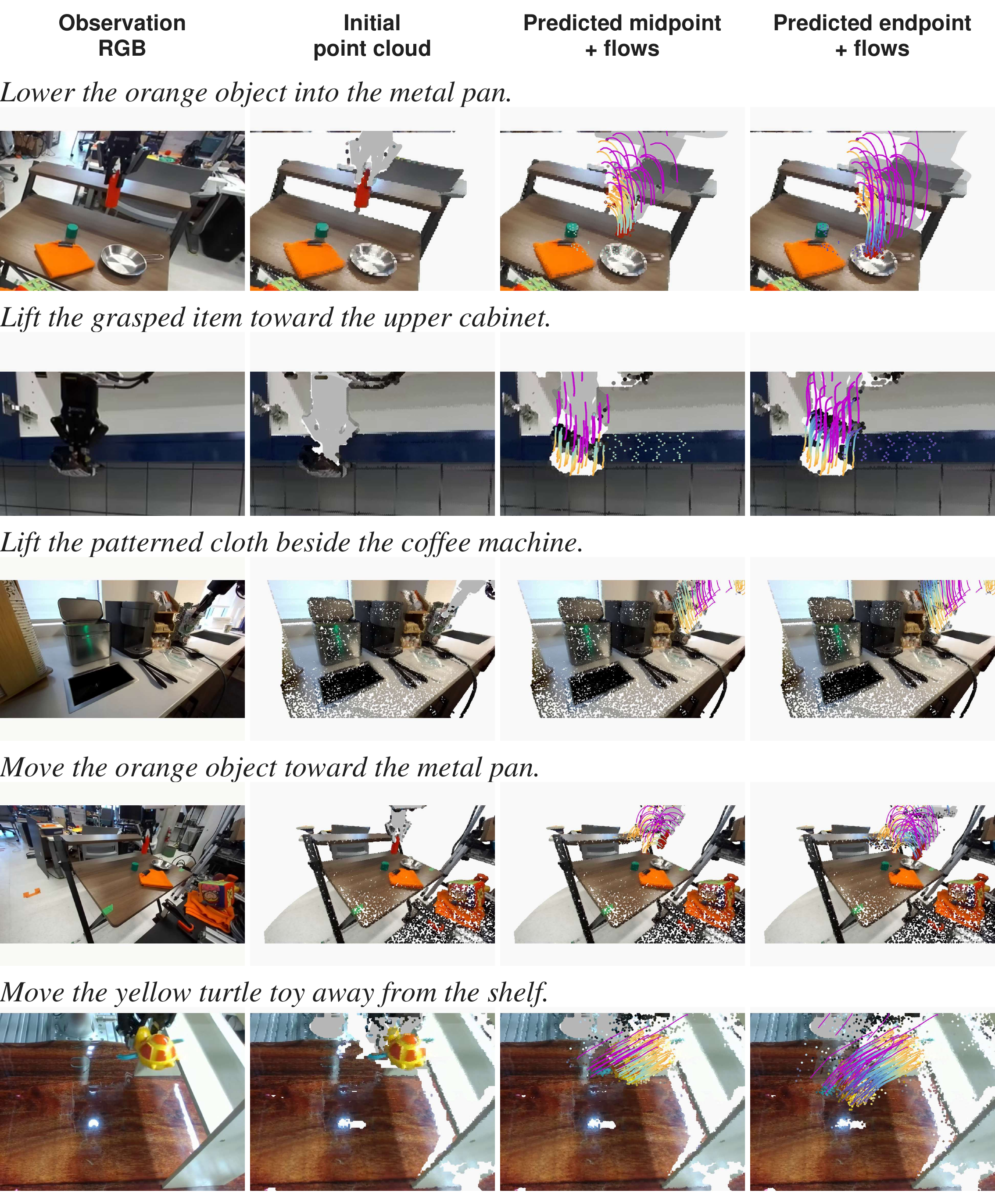}
    \caption{\textbf{Zero-shot scene-flow predictions on real-world observations.}
    The first four rows are from DROID and the last is from RH20T.
    Columns show observed RGB, initial points, and predicted midpoint and
    endpoint point clouds with scene trajectories (orange to purple) and
    ground-truth gripper trajectories used for conditioning (magenta).
    No scene-trajectory ground truth is shown.}
    \label{fig:app:real_scene_flow}
\end{minipage}
\end{center}

\end{document}